\pdfoutput=1

\documentclass[11pt,table]{article}

 \usepackage[]{acl}

\usepackage{times}
\usepackage{latexsym}
\usepackage{pifont}

\usepackage{tikz}
\usetikzlibrary{positioning}
\usepackage[T1]{fontenc}

\usepackage[utf8]{inputenc}
\usepackage{breqn}
\usepackage{graphicx}
\usepackage{amssymb}
\usepackage{amsmath}
\usepackage{algorithm}
\usepackage{microtype}
\usepackage{multirow}
\usepackage{url}
\usepackage{amsmath}
\usepackage[textsize=scriptsize]{todonotes}
\usepackage{booktabs}
\usepackage{comment}
\usepackage{enumitem}
\usepackage{tabu}
\usepackage{dsfont}
\usepackage{algpseudocode}
\usepackage{listings}

\usepackage{amsthm}
\newtheorem{theorem}{Theorem}[section]

\newtheorem{define}{Definition}[section]

\definecolor{codegreen}{rgb}{0,0.6,0}
\definecolor{codegray}{rgb}{0.5,0.5,0.5}
\definecolor{codepurple}{rgb}{0.58,0,0.82}
\definecolor{backcolour}{rgb}{0.95,0.95,0.92}

\lstdefinestyle{mystyle}{
    backgroundcolor=\color{backcolour},   
    commentstyle=\color{codegreen},
    keywordstyle=\color{magenta},
    numberstyle=\tiny\color{codegray},
    stringstyle=\color{codepurple},
    basicstyle=\ttfamily\footnotesize,
    breakatwhitespace=false,         
    breaklines=true,                 
    captionpos=b,                    
    keepspaces=true,                 
    numbers=left,                    
    numbersep=5pt,                  
    showspaces=false,                
    showstringspaces=false,
    showtabs=false,                  
    tabsize=2
}

\newcommand\pythonstyle{\lstset{
language=Python,
basicstyle=\ttm,
otherkeywords={self},             
keywordstyle=\ttb\color{deepblue},
emph={MyClass,__init__},          
emphstyle=\ttb\color{deepred},    
stringstyle=\color{deepgreen},
commentstyle=\color{deepred}\sffamily,
frame=tb,                         
showstringspaces=false,            %
mathescape=true
}}

\lstnewenvironment{python}[1][]
{
\pythonstyle
\lstset{#1}
}
{}

\newcommand\pythoninline[1]{{\pythonstyle\lstinline!#1!}}

\usepackage{multirow}
\usepackage{bbm}

\newcommand{\greedy}{\textsc{Greedy}}
\newcommand{\bs}{\textsc{BS}}
\newcommand{\dbs}{\textsc{DBS}}
\newcommand{\topp}[1]{\textsc{Ncls}$_{#1}$}
\newcommand{\temp}[1]{\textsc{Temp}$_{#1}$}

\newcommand{\bfs}{\textsc{Bfs}}

\newcommand{\base}{\textsc{Rcb}}
\newcommand{\zip}{\textsc{Zip}}

\newcommand{\zhang}{\textsc{ZBeam}}

\newcommand{\nodesos}{$n_{\textrm{sos}}$}
\newcommand{\nodeeos}{$n_{\textrm{eos}}$}
\newcommand{\ebs}{$k$}

\newcommand{\gen}{\textsc{Gen}}
\newcommand{\mrg}{\textsc{Mrg}}

\newcommand{\numpath}{$|$path$|$}
\newcommand{\novone}{\textsc{N}1}
\newcommand{\novtwo}{\textsc{N}2}
\newcommand{\selfbleu}{\textsc{sBl}}
\newcommand{\edit}{\textsc{ED}}

\newcommand{\tabgram}{\textsc{Grm}}
\newcommand{\errate}{\textsc{Err}}
\newcommand{\bleu}{\textsc{Bl}}

\newcommand{\taboracle}{\textsc{Or}}

\newcommand{\tabsample}{\textsc{Sp}}

\definecolor{gcolor}{RGB}{42, 157, 143}
\definecolor{bcolor}{RGB}{231, 111, 81}

\newcommand{\badcell}{\cellcolor{BrickRed!15}}
\newcommand{\finecell}{\cellcolor{RoyalBlue!10}}
\newcommand{\goodcell}{\cellcolor{RoyalBlue!30}}

\newcommand{\pathh}{\mathbf{h}}
\newcommand{\candh}{\mathbf{\hat{h}}}

\DeclareMathOperator*{\argmax}{arg\,max}

\renewcommand{\algorithmiccomment}[1]{\bgroup\hfill\small//~#1\egroup}

\title{Massive-scale Decoding for Text Generation using Lattices}

\author{Jiacheng Xu\textsuperscript{\ding{117}} \quad Siddhartha Reddy Jonnalagadda\textsuperscript{\ding{72}} \quad Greg Durrett\textsuperscript{\ding{117}}\\
\textsuperscript{\ding{117}}Department of Computer Science, The University of Texas at Austin \\
\textsuperscript{\ding{72}}Alexa AI, Amazon, Seattle \\
  \texttt{\{jcxu,gdurrett\}@cs.utexas.edu}  \quad \texttt{sidjreddy@gmail.com}\\}

\begin{document}
\maketitle
\begin{abstract}
Conditional neural text generation models generate high-quality outputs, but often concentrate around a mode when what we really want is a diverse set of options. 
We present a search algorithm to construct lattices encoding a massive number of generation options. 
First, we restructure decoding as a \textit{best-first search}, which explores the space differently than beam search and improves efficiency by avoiding pruning paths. 
Second, we revisit the idea of \textit{hypothesis recombination}: we can identify pairs of similar generation candidates during search and merge them as an approximation. 
On both summarization and machine translation, we show that our algorithm encodes thousands of diverse options that remain grammatical and high-quality into one lattice. 
This algorithm provides a foundation for building downstream generation applications on top of massive-scale diverse outputs.\footnote{Code, implementation guideline, and visualization are available at \url{https://github.com/jiacheng-xu/lattice-generation}. }

\end{abstract}

\section{Introduction}


Although pre-trained text generation models 
\cite{lewis-etal-2020-bart,RaffelEtAl2020} have achieved impressive results across a range of tasks, these models do not always deliver what system developers want. Machine generated text may be non-factual \cite{kryscinski-etal-2020-evaluating,maynez-etal-2020-faithfulness,goyal-durrett-2021-annotating} or toxic  \cite{gehman-etal-2020-realtoxicityprompts}. 
We might patch these problems by applying discriminators over the output \cite{holtzman-etal-2018-learning,yang-klein-2021-fudge} to enforce these properties post-hoc; we could, for instance, apply a secondary model as a reranker over a small collection of outputs. However, if the generator returns a homogeneous set of candidates, we may fail to find \emph{any} usable generation output.
What if generation models could return \textit{massive} numbers of candidates rather than a few outputs with optimal score?
With a large set of candidates, our secondary model could more easily find an acceptable one without having to take more extreme steps like re-training the initial generation model.
Output diversity has separately been established as a useful goal for for applications such as dialogue and story generation \cite{li-etal-2016-diversity,fan-etal-2019-strategies}. 

Standard approaches including beam search (\bs) and sampling methods fall short of our goal. Beam search uses significant computational resources to explore similar hypotheses, and much of the computation in the search process is invested into paths that could be acceptable generation outputs, but are ultimately pruned. Sampling approaches like nucleus sampling \cite{Holtzman2020The}, although achieving better diversity than beam search, often re-discover seen hypotheses and can be harder to control for quality.
A central problem with both methods is that they do not handle very similar hypotheses efficiently.


\begin{figure}[t]
    \centering
    \includegraphics[width=0.50\textwidth]{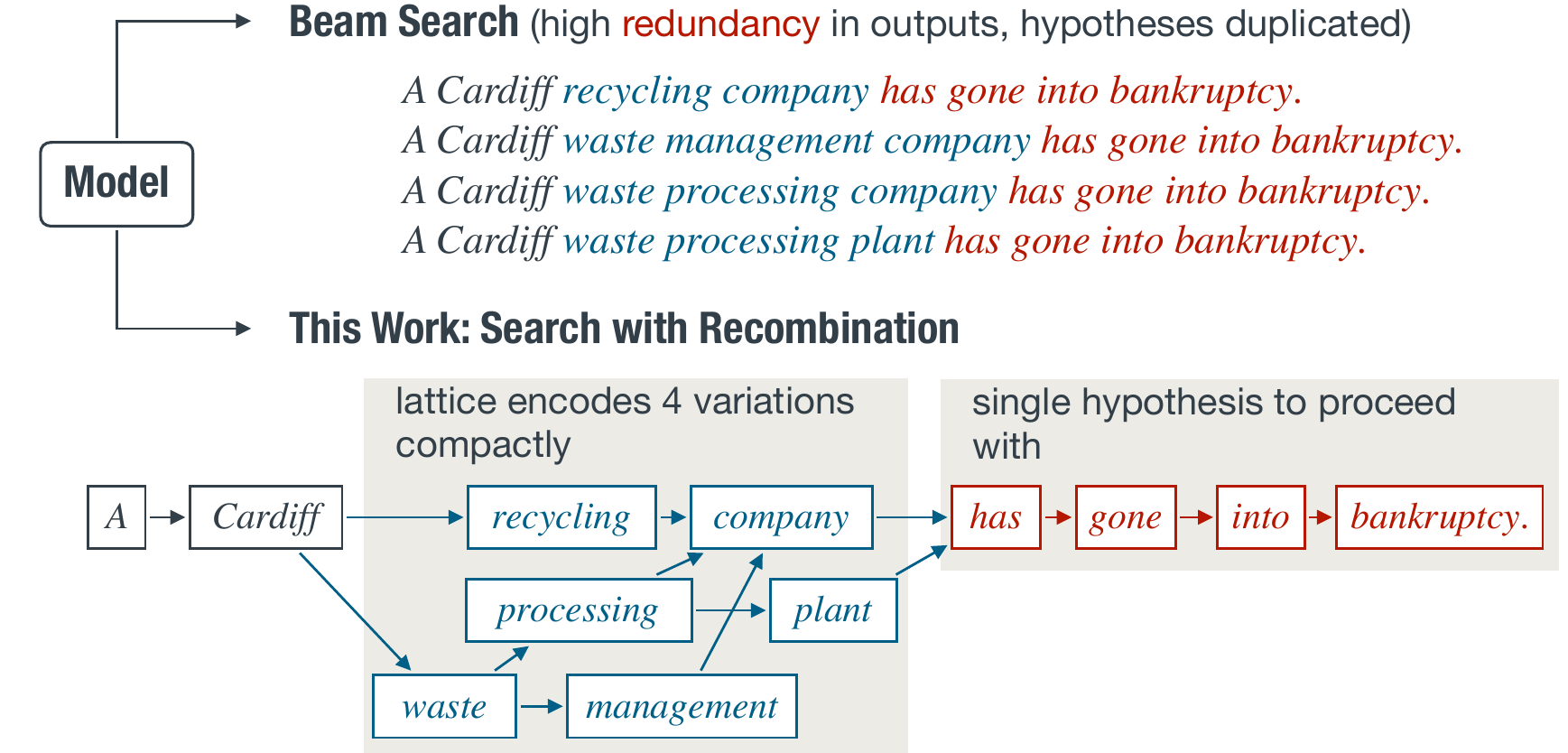}
    \caption{A lattice of outputs yielded by path recombination is a more efficient way to represent and explore related generation outputs compared to beam search.}
    \label{fig:beamex}
\end{figure}

In this paper, we present a decoding framework with two key components. 
First, we argue that a modified \textbf{best-first search} (\bfs) is the right way to explore the search space. We augment standard best-first search with a depth-first path completion strategy: we eagerly expand each node until we reach an EOS token, thereby guaranteeing that each node is part of some completed path returned to the user. This generation strategy avoids exploring large numbers of states which end up being pruned. \bfs{} is also more flexible than static beam search and can prioritize exploration in more uncertain parts of the generation.

\begin{figure}[t]
    \centering
    \footnotesize
    \includegraphics[width=0.482\textwidth]{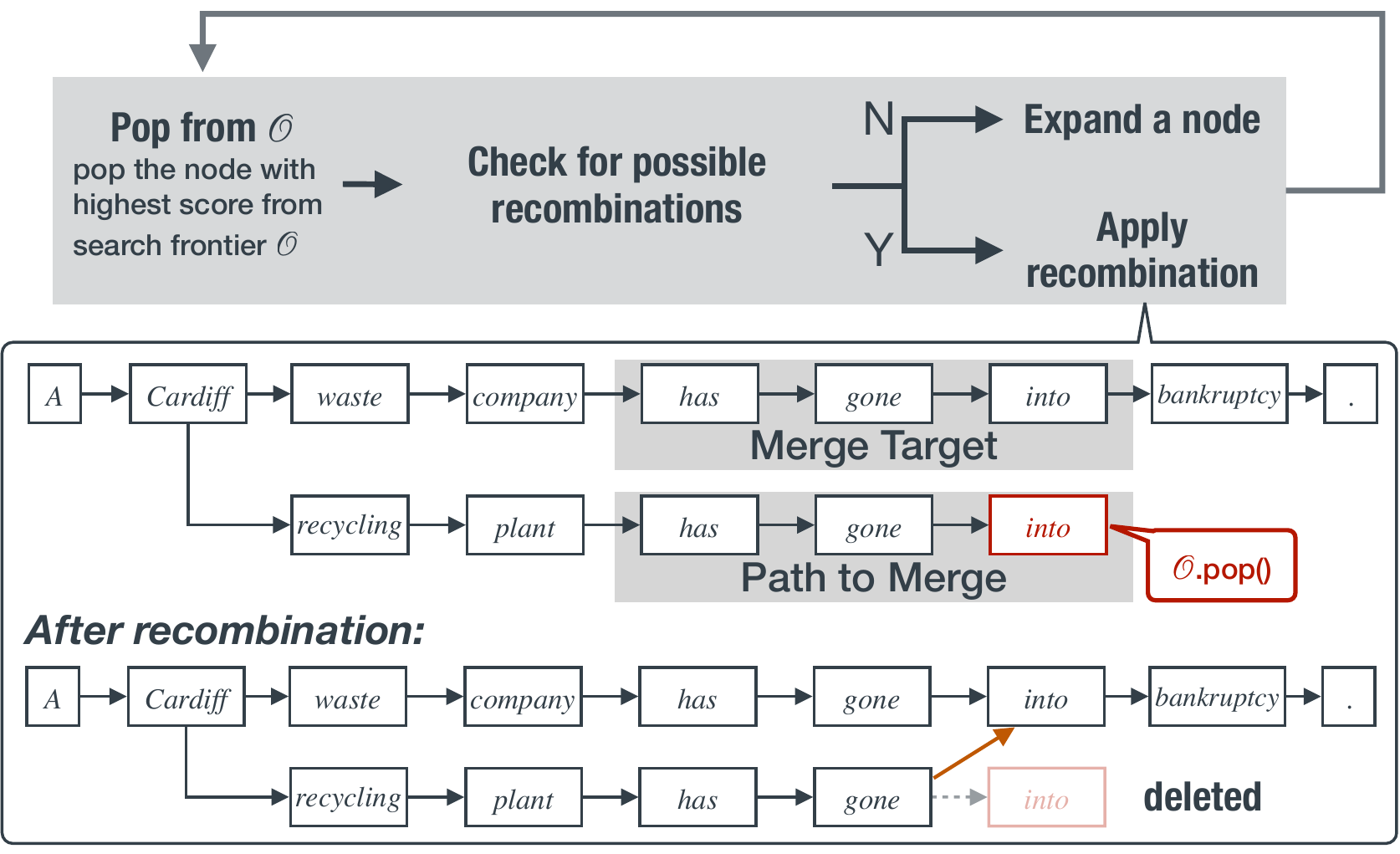}
    \caption{Our search algorithm. At each step, the algorithm pops a node from search frontier $\mathcal{O}$, checks for possible recombinations with existing nodes, and merges the nodes if a match is found. In this example, ``\emph{has gone into}'' is the merge target to match. ``\emph{waste company}'' and ``\emph{recycling plant}'' are interchangeable paraphrases which do not affect the continuation from the model's perspective.  }
    \label{fig:searchintro}
\end{figure}

Second, our algorithm returns a massive number of generation options encoded in a lattice, with different hypotheses \textbf{recombined} in an approximate fashion. Beam search preserves similar outputs such as ``\emph{A Cardiff recycling company has gone into}'' and ``\emph{A Cardiff waste management company has gone into}'' as different states. However, these prefixes actually have very similar distributions of following words under the model; if we identify states like this, we can recombine them (Figure~\ref{fig:searchintro}) and treat them as the same from the perspective of future continuations. 
In Figure~\ref{fig:beamex}, we show an illustration of the lattice structure this recombination can form for document summarization. We broaden a recombination method used previously in beam search for machine translation \cite{och-etal-2001-efficient,zhang-etal-2018-exploring}, enabling us to compactly encode large number of generation candidates and achieve dense lattices.

We show results for both document summarization and machine translation in three language pairs. For each setting, we show that our lattice encodes a large number of high-quality candidates, including good matches with annotated reference generations. We further show that a variant of our method can still achieve strong results with a lower number of nodes expanded than the baselines, suggesting that this can be a path towards saving computational resources. We believe that computing thousands of high-quality generation candidates within a single compact data structure can provide a powerful starting point for various downstream purposes: diversity, factuality, customizability, and more.

\section{Problem \& Setup}

We define our algorithm in the context of conditional text generation \cite{sutskever2014sequence,bahdanau2014neural}. Conditional text generation is formulated as sequence transformation from a source input $\mathbf{x}$ to target output $\mathbf{y} = (y_1,\ldots,y_n)$ via a neural text generation model parameterized by $\theta$. Each $y_i$ is a symbol in a vocabulary $\mathcal{V}$.
The probability of a decoded sequence is given by $p(\mathbf{y} \mid \mathbf{x}; \theta) = \prod_{t=1}^n p(y_{t} \mid \mathbf{y}_{<t}, \mathbf{x}; \theta)$. 
Decoding text from a model can be framed as a search problem, where the search objective is to find the output sequence that maximizes the conditional probability under the model: $\argmax_{\hat{\mathbf{y}}} p(\hat{\mathbf{y}} \mid \mathbf{x}; \theta)$.
Because $p(\hat{y}_{t} \mid \hat{\mathbf{y}}_{<t}, \mathbf{x}; \theta)$ depends on the entire generated sequence so far, this decoding problem is intractable to solve exactly.

While typically the goal of decoding is to find the hypothesis with the highest possible model score, we instead focus on finding a large set of ``good enough'' hypotheses.
That is, finding a set $\mathcal{Y}$:
\begin{equation}
\label{eqn:goal}
\small
    \argmax_{\mathcal{Y}}\ \ |\mathcal{Y}| \textrm{\quad s.t. }\   p(\mathbf{y} \mid \mathbf{x}; \theta) > \epsilon\ \textrm{for all}\  \mathbf{y} \in \mathcal{Y}
\end{equation} 
for some threshold $\epsilon$. $\epsilon$ emerges naturally by adjusting search hyperparameters to control the number of returned hypotheses. Our goal in this paper is to design an algorithm that can efficiently find $\mathcal{Y}$.



\paragraph{Notation}

We encode predicted generation candidates $\hat{\mathbf{y}}$ in a lattice. A \textbf{lattice} $L = (N, E)$ is a directed graph where each \textbf{node} represents a token and paths defined by directed edges encode candidates. A \textbf{path} $\pi$ in $L$ from a unique start-of-sequence node \nodesos{} to any node $n$ represents a (partially) decoded string, consisting of the words in that path. All completed paths start with \nodesos{} and end at (potentially different) end-of-sequence nodes \nodeeos.
The search graph $L$ is constructed iteratively through a search procedure. We maintain the closed graph $\mathcal{C}$ with explored nodes and edges as well as a search frontier $\mathcal{O}$, a set consisting of successors to nodes currently in the graph. For each node, there are $|\mathcal{V}|$ possible successors.

We define the \textbf{search budget} as the number of nodes expanded from the search frontier. Our experiments will seek to compare different methods using the same search budget. We will define this more precisely in Sec.~\ref{sec:evaluation}.



\section{Inadequacies of Beam Search}
\label{sec-beam-fail}
As we have alluded to, beam search is inadequate for our goal for several reasons. We show experiments on these aspects in this section and Appendix~\ref{app-beam-fail}.

\begin{figure}[t]
    \centering
    \includegraphics[width=0.4825\textwidth]{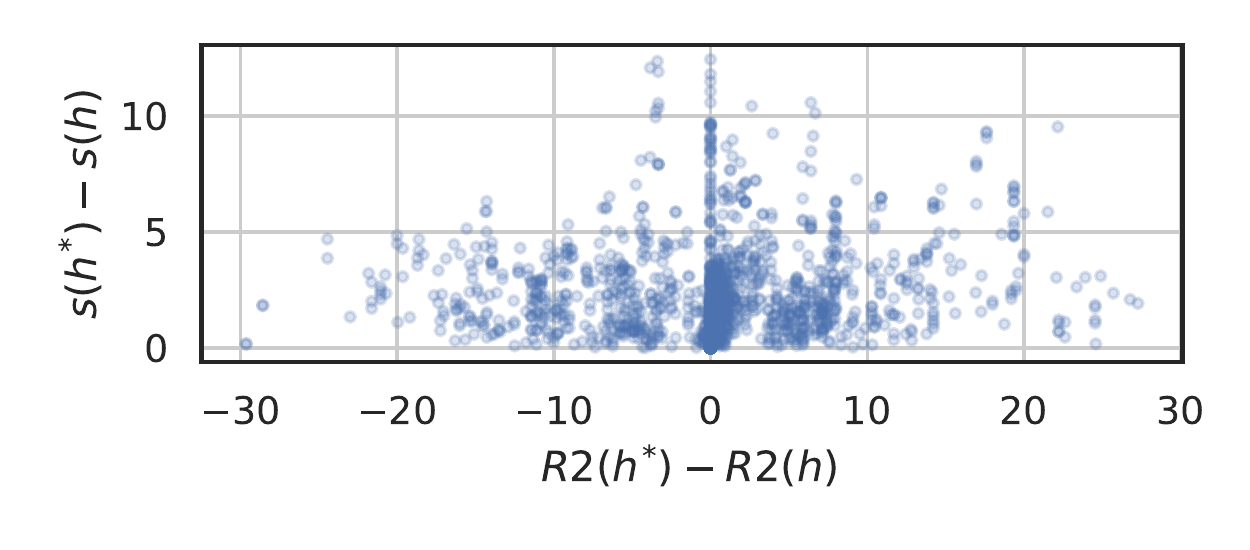}
    \caption{Correlation of ROUGE-2 and model score in beam search. For each example, we compare the hypothesis with the highest model score, $h^{*}$, with all other hypotheses.  $x$ and $y$-axis show the gaps of R2 and model score. The Pearson's $\rho$ is 0.092 which suggests very low correlation between R2 and model score.}
    \label{fig:corr}
\end{figure}

\paragraph{Better Model Score $\nRightarrow$ Better Hypothesis }
\label{sec:wrong_goal}
The most critical issue is that beam search is designed to efficiently approximate $\argmax \hat{\mathbf{y}} = p(\hat{\mathbf{y}} \mid \mathbf{x}; \theta)$, but the optimal model score is neither our goal nor a guarantee of a good hypothesis.
In Figure~\ref{fig:corr}, we compare the correlation of model score and ROUGE under beam search for text summarization. The Pearson correlation between these two variables is very weak. Beyond ROUGE score, the example in Fig.~\ref{fig:beamex} shows that the main differences between these summaries may be minor differences in surface realization that have little effect on our qualitative judgments of summary quality.
\textbf{The hypothesis with the best score under beam search is not substantially better quality than hypotheses with slightly lower scores.} 

\begin{figure}[t]
    \centering
    \footnotesize
    \includegraphics[width=0.4825\textwidth]{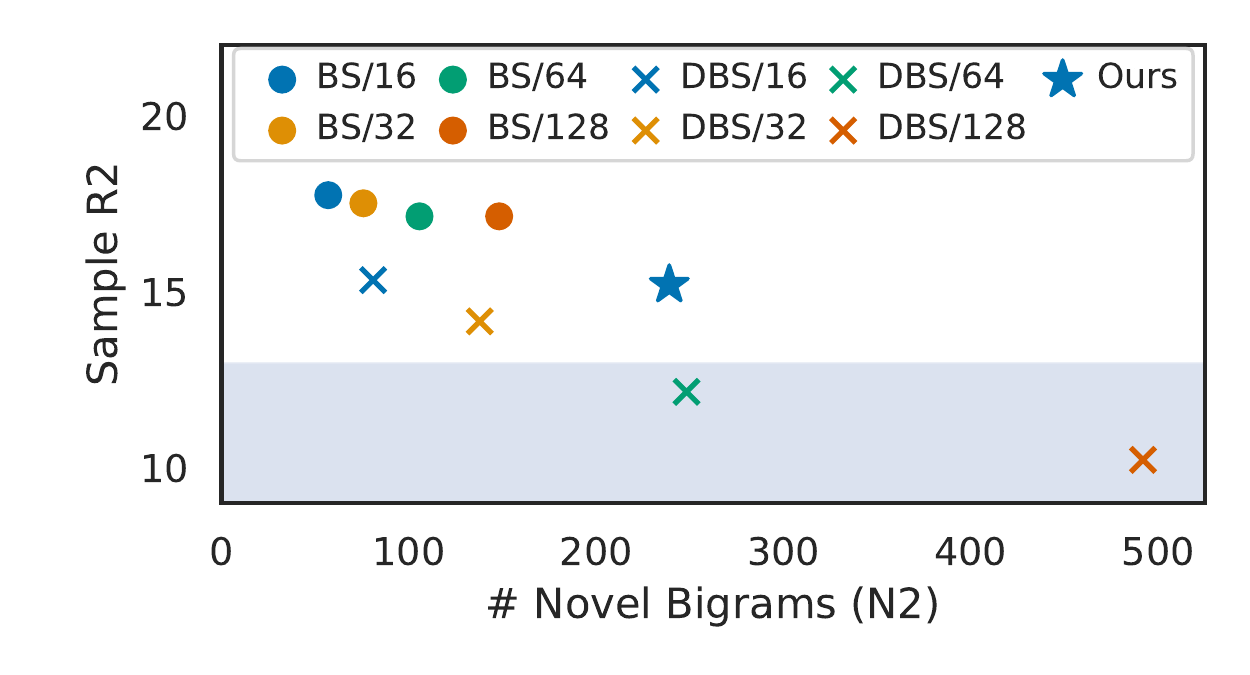}
    \caption{Diversity vs.~ROUGE of \bs{}/\dbs{} on XSum with varying beam size $k$, compared to a proposed model introduced later (blue star) with equivalent beam size $k=16$. We consider sample ROUGE-2 lower than 13 as low relevance/quality generations.
    Diversity of \bs{} does not scale well with $k$ and \dbs{} generations become low quality. }
    \label{fig:large-beam-tradeoff}
\end{figure}

\paragraph{Lack of Diversity in (Diverse) Beam Search}
Are the model outputs from \bs{} and \dbs{} diverse? We use Self-BLEU (\selfbleu) \cite{zhu2018texygen} to measure the BLEU score for randomly sampled pairs from each algorithm's output. The lower the self-BLEU, the more dissimilar the pairs are. 
On decoding summaries on XSum, the \selfbleu{} for \bs{}/\dbs{} are 87/79 while a nucleus sampling method can achieve 57/50 depending on the configuration.  
Although \dbs{} slightly improves the diversity compared to the original variant, \textbf{the overlap of outputs from beam search based method is still very high, and the diversity remains a challenge.}

\begin{table}[t]
\centering
\footnotesize
\begin{tabular}{@{}r|c|ccc@{}}
\toprule
$k$ & 16     & \multicolumn{3}{c}{8}    \\ 
$\mathcal{D}$ & XSum   & zh-en  & fr-en  & en-fr  \\ \midrule
\bs   & 71.3\% & 63.3\% & 54.0\% & 59.2\% \\
\dbs  & 71.2\% & 56.1\% & 50.4\% & 55.7\% \\ \bottomrule
\end{tabular}
\caption{Pruning ratio of \bs{} and \dbs{} on different tasks and datasets with beam size $k$. We report the average percentage of explored nodes getting pruned and not appearing in a finished hypothesis.} 
\label{tab:prune}
\end{table}

\paragraph{Poor Efficiency from Pruning}

One final issue with beam search is that \textbf{most of its computation is not even useful in producing finished hypotheses}; that is, the set $\mathcal{Y}$ of answers produced does not contain most of the nodes expanded in the typical course of operation. 
We conduct an empirical pruning study on a summarization dataset and three translation datasets and show the results in Table~\ref{tab:prune}. For all studied cases, beam search and diverse beam search prune over half of the expanded nodes. 
Many pruned hypotheses are not truly ungrammatical or low quality, but are merely slightly less likely than other nodes. How we can preserve more of the explored lattice and do so efficiently is addressed in next by our use of best-first search. 


\begin{algorithm}[t]
\small
\caption{Best-first search with depth-first completion and path recombination}
\label{algo-bfs}
\begin{algorithmic}[1]
\Require Generation model $\theta$ with vocabulary $\mathcal{V}$, search budget $b$, $\mathcal{O}$ and $\mathcal{C}$ denote open set (max priority queue) and closed set, $isRecomb$ and $doRecomb$ are functions checking and running path recombination. 
\Ensure All completed paths $P$ 
\State $\mathcal{O} \leftarrow$ \{($\infty$,\nodesos{})\}, $\mathcal{C}\leftarrow \emptyset$, $expanded \leftarrow 0$.
 \While{ $expanded < b$ }
 \State $ \hat{h} \leftarrow \mathcal{O}$\texttt{.pop()}
   \If{ $isRecomb(\hat{h}, \mathcal{C})$}
    \State $doRecomb(\hat{h}, \mathcal{C})$
    \State \textbf{continue}
  \EndIf
\If{$\hat{h}  \neq \text{EOS} $}
\State $ v_{greedy} = \argmax_{v \in \mathcal{V}}  p(v \mid  \hat{h}, \mathbf{x}; \theta )$
\For{ $v \in \mathcal{V}$}
\State score $\leftarrow s( \hat{h} \bigoplus v) $ \Comment{concatenation}
\If{ $v = v_{greedy}$}
\State score $\leftarrow \infty $ \Comment{depth-first completion}
\EndIf
\State  $\mathcal{O} \leftarrow  \mathcal{O} \cup (\text{score}, n_{v} )$ 
\EndFor
\State    $expanded \leftarrow expanded + 1$
\EndIf
\State $ \mathcal{C} \leftarrow \mathcal{C} \cup \hat{h}$
\EndWhile
\end{algorithmic}
\end{algorithm}

\section{Modified Best-first Search}
\label{sec-bfs}
As established in the previous section, beam search prunes many paths that would potentially yield high-quality summaries and wastes computational resources expanding nodes that aren't included in a final search graph.
We tackle this issue by changing from beam search to \emph{best-first search} (\bfs) \cite{hart1968formal,pearl1984heuristics}. \bfs{} prioritizes searching over nodes according to a scoring function, giving us more flexibility in how we explore the space.
Our chief modification of the base algorithm is a heuristic we call depth-first completion. 

\paragraph{Depth-first Path Completion}
Neural text generation is a search problem with large branching factor ($\mathcal{V}$) and deep search depth (sequence length). As a result, applying \bfs{} with the scoring function being the model score of a state often leads to a broad search that rarely returns a valid path. 
One solution to this problem is to incorporate a heuristic based on length. Model score is monotonically decreasing as a sequence grows in length, so prior work \cite{wu2016google,zhang-etal-2018-exploring,meister-etal-2020-best} has used a length reward term to alleviate this issue.\footnote{This can be considered a heuristic like in (weighted) A$^*$ search, but it is not necessarily admissible or consistent.} 
We found that, even with a length heuristic, \bfs{} will still have ``dangling'' nodes that are not part of any path to an EOS (goal) token, and it might return few or no valid hypotheses.

\begin{figure}[t]
    \centering
    \footnotesize
    \includegraphics[width=0.4825\textwidth]{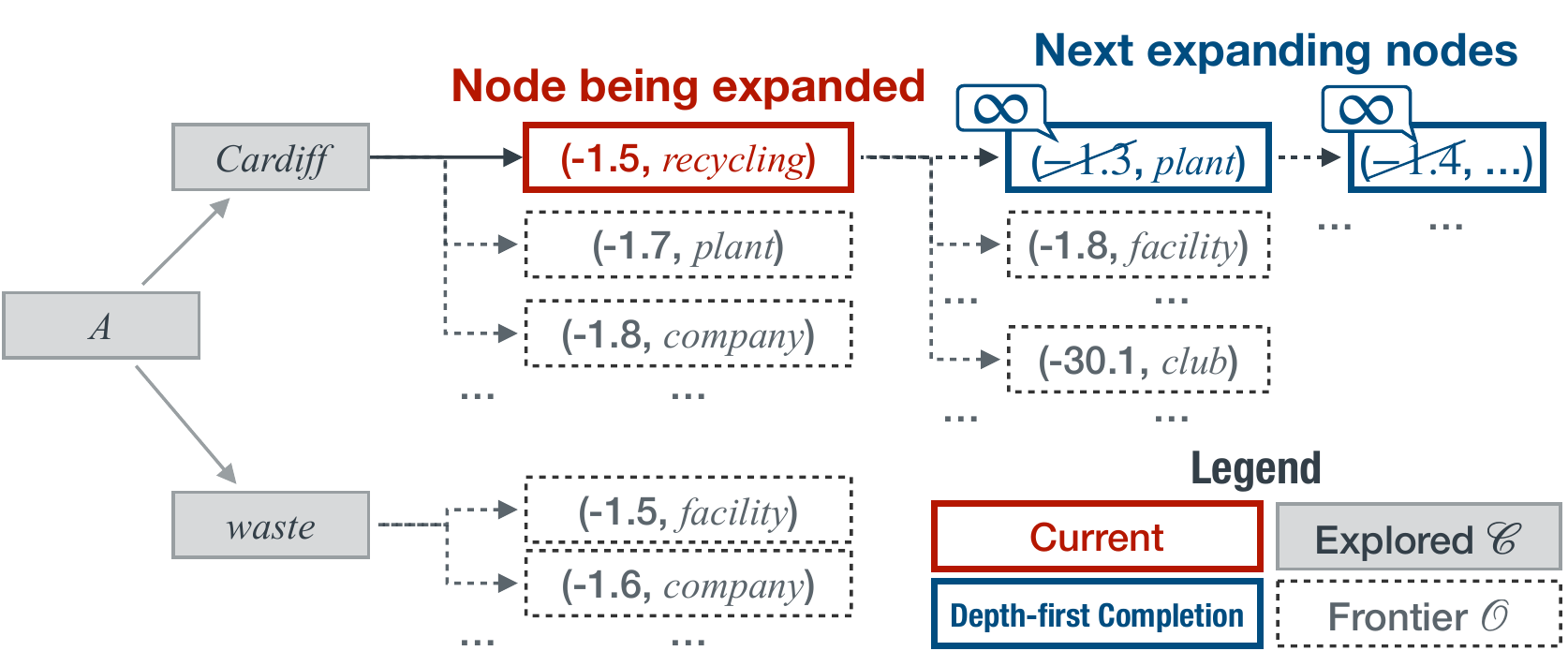}
    \caption{ Depth-first completion. The red node is the current node being expanded. We depth-first expand a sequence of nodes (in blue) to get a completed path.   }
    \label{fig:rollout}
\end{figure}

Recognizing our objective from Equation~\ref{eqn:goal}, we can take a simple step to ensure that every node ends up on some completed path: \textit{eagerly do a greedy search from each node until we reach \nodeeos{} or exceed a maximum length}. In Algorithm~\ref{algo-bfs}, we implement this by modifying the priority of the highest scored token with $\infty$ (line 12), so it will be explored immediately after the current time step. In Figure~\ref{fig:rollout}, we illustrate depth-first completion.

\paragraph{Search Algorithm}
We describe \bfs{} with depth-first completion in Algorithm~\ref{algo-bfs}. The algorithm is a modified best-first search algorithm applied to text generation. $s(\cdot)$ is a function to evaluate the value of a path. Typically it is defined as $s(\mathbf{y}) = \sum \log p(y_t \mid y_{<t})$. $b$ is the budget for total model calls to neural text generation model. Note that $isRecomb$ and $doRecomb$ do not invoke the neural generation model, so they do not count towards the computation budget we defined here. In practice, we only consider top 5 expansions rather than the whole vocabulary $\mathcal{V}$ for line 10.

\section{Path Recombination}
\label{sec_rec}

Path recombination, also known as hypothesis recombination, was originally proposed and used in phrase-based machine translation \cite{och-etal-2001-efficient,koehn-etal-2003-statistical,zhang-etal-2018-exploring}. 
The idea of path recombination is to \emph{combine similar paths if what the model predicts for them in the future is the same}, 
reflecting a similar dynamic programming principle as the Viterbi algorithm. We focus on finding hypotheses which approximately exhibit this property, and show that merging them can yield high-quality outputs.
Figure~\ref{fig:searchintro} shows an example of recombination. The two hypotheses being merged here roughly convey the same intent, and it turns out that the shared suffix ``\emph{has gone into}'' is a strong indicator that the model will treat them similarly in the rest of the generation.



\paragraph{Prerequisites of Recombination}
Theoretically, two search states should only be recombined if they yield the exact same distribution over future generation decisions (see strong equivalence in Appendix~\ref{app:strong}). However, this is intractable even to check approximately; we define a weaker criterion:
\begin{define}[Weak equivalence]
\label{assump-weak}
Let $\mathbf{a}$ and $\mathbf{b}$ be two prefix strings starting with \nodesos{}.
$\mathbf{a}$ and $\mathbf{b}$ are weakly equivalent if greedy completions of these two strings are the same: $\argmax_{\mathbf{y}} P(\mathbf{y} \mid \mathbf{a}) = \argmax_{\mathbf{y'}} P(\mathbf{y'} \mid \mathbf{b})$.
\end{define}
\noindent
This criterion can be checked empirically, but it is not practical to do so during search. 

\begin{figure}[t]
    \centering
    \footnotesize
    \includegraphics[width=0.4825\textwidth]{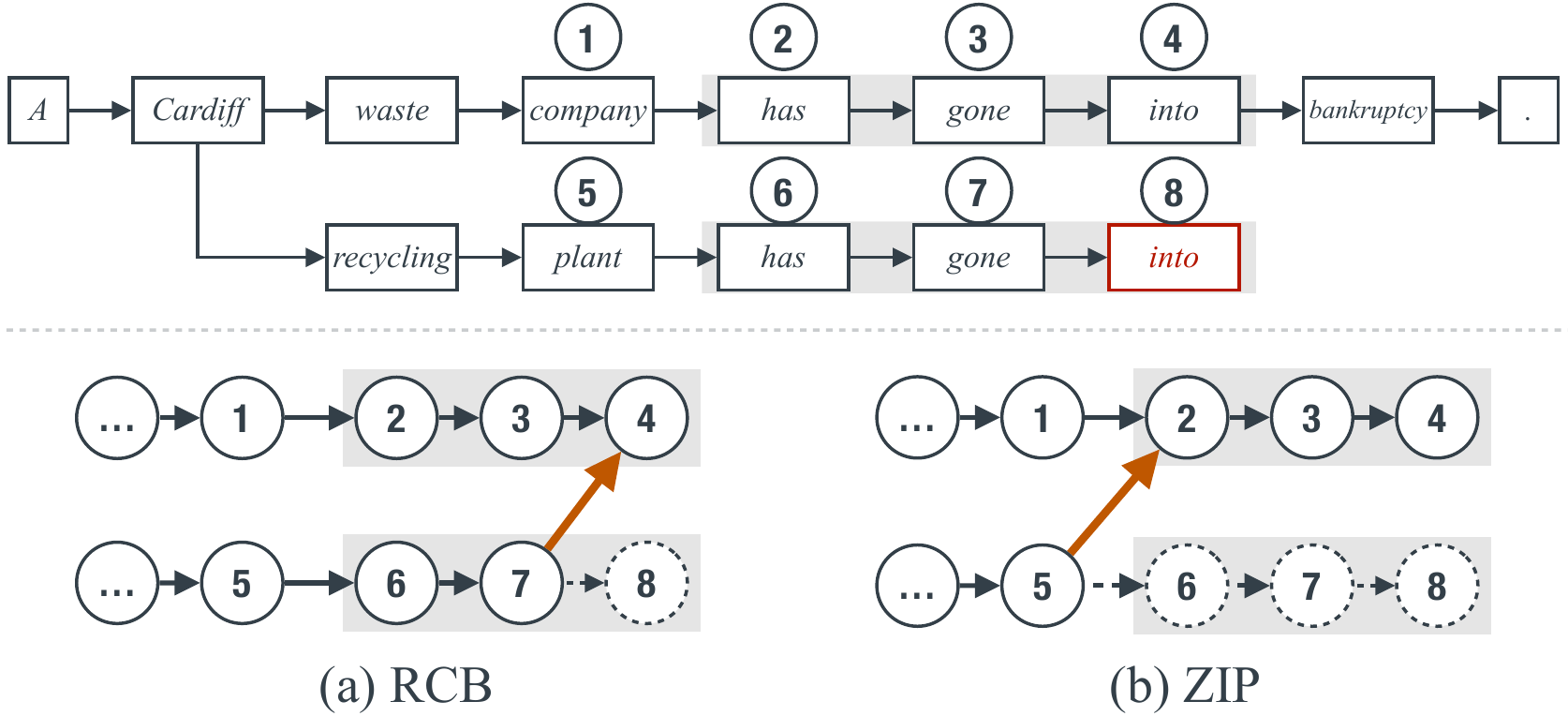}
    \caption{Illustration of two path recombination strategies, \base{} and \zip{}. 
    Orange lines are the merging edges (\mrg{}) built by recombination. Dotted lines and circles are removed after recombination. The key difference of \base{} and \zip{} is how much the recombination propagates, 1 step or $n$ steps. }
    \label{fig:fullrecomb}
\end{figure}



To approximate equivalence, we define a similarity function $\textit{isRecomb}(\pathh, \candh)$ to determine if an expanded node $\candh$ should be merged with an existing expanded node $\pathh$.
A similar recombination idea was explored in \citet{zhang-etal-2018-exploring}. Following their work, we explore a family of rule-based heuristics for merging. There are two rules: (1) two strings share a common $n$-gram suffix, (2) the length difference of two strings is less than $\alpha$. Assume that the canonical paths for $\pathh$ and $\candh$ are lengths $l$ and $\hat{l}$, then
$ \textit{isRecomb}(\pathh,\candh ) = \mathbbm{1}[\pi(\pathh)_{l-n+1,\ldots,l}  = \pi(\candh)_{\hat{l}-n+1,\ldots,\hat{l}}  \wedge | l - \hat{l} | < \alpha]$
where $\alpha$ and $n$ are hyper-parameters.\footnote{In \citet{zhang-etal-2018-exploring}, there is one extra constraint requiring $P(\candh \mid \mathbf{x}) < P(\pathh \mid \mathbf{x})$, which requires that the path getting recombined has lower model score than the existing path. However, we found that model score is not always a good indicator for merging, as suggested in Fig.~\ref{fig:corr}, partially because it is challenging to calibrate scores across different sequence lengths, so we disregard this constraint.}
For a large enough value of $n$, note that the shared suffixes encourage hypotheses like this in Figure~\ref{fig:fullrecomb} that share large parts of the structure already.

\paragraph{Prior Work: \bs\zhang}
\citet{zhang-etal-2018-exploring} use their merging criterion in the context of beam search for neural machine translation. 
If the merging criteria hold, $\candh$ will be recombined with $\pathh$. However, $\candh$ will not be considered as a future merging candidate.
We call this merging strategy \zhang. We implement this model together with its merging criteria and denote it as \bs\zhang. This strategy is tailored to beam search and explores a more limited set of merges than we do. 

\paragraph{Canonical Paths}
After recombination, a single node may represent multiple different possible sentence prefixes. If an edge is created due to the extension of search graph via model's prediction, we call it a \gen{} edge. Otherwise, the edge is created due to path recombination, and we call it a \mrg{} edge. 
We define the notion of a canonical path, which represents the single path used to score candidate expansions.

\begin{define}[Canonical Path]
\label{define-init}
Let $n$ be a node. The canonical path to $n$ is defined as the unique path from \nodesos{} to $n$ consisting only of \gen{} edges.
\end{define}

\begin{theorem}[]
\label{theorem:can}
For any node $n$ in the graph except \nodesos{}, there exists exactly one canonical path.
\end{theorem}
\noindent
We present the proof in Appendix.~\ref{app:proof}.
We define the path of a node $n$, $\pi(n)$, as returning the sequence of words corresponding to the canonical path of that node. Expanding $n$ computes $P(y \mid \pi(n))$ under the neural model.

\section{Recombination Mechanism}

We illustrate the two major recombination techniques we introduce, \base{} and \zip{}, in Figure~\ref{fig:fullrecomb}. These methods represent our two implementations of \emph{doRecomb} in Algorithm~\ref{algo-bfs}.







\paragraph{\base: Generalization of \zhang}
\zhang{} has a major limitation: a limited set of merging candidates.
The potential merge candidates in \zhang{} are only nodes in the current beam hypotheses and their previous steps, so the method cannot merge with nodes from earlier timesteps. For example, ``\emph{A waste plant has gone into}'' cannot be merged with the hypothesis with ending in node 4 shown in Figure~\ref{fig:fullrecomb}. 
The proposed generalization, \base{}, addresses this limitation. 
We index all of the nodes in the lattice across all timesteps by their $n$-grams using a hash table, making it $O(1)$ time to look up an $n$-gram pattern and retrieve potential merge candidates if they exist. 


\paragraph{\zip{}: Recombining More} 

If we take a closer look at \base{} in Figure~\ref{fig:fullrecomb}, we see that even in the merged structure, nodes 3 and 7 and nodes 2 and 6 are preserved as separate. They do not pass the recombination criterion themselves, but these nodes are part of the suffix matched strings, still correspond to the same words, and have the same directly generated next word. There is reason to believe that these might be equivalent as well.
Hence, we explore a variant called \zip{} that propagates the merge backwards through the lattice. 
This change relaxes the merging criterion and up to $n$ pairs of nodes are combined when a merge is identified, leading to a more compact lattice.
We describe some of the details in Appendix~\ref{app:zip}. 

\paragraph{Our Methods}
In this work, we combine the two proposed techniques, the modified best-first search (\bfs{}) and recombination methods, together. Hence, we name them as \bfs{}\base{} and \bfs{}\zip{}. 




\section{Evaluation}
\label{sec:evaluation}

To evaluate the proposed methods, we conduct experiments on abstractive text summarization and machine translation. Our evaluation focuses on two questions: (1) how \textbf{large and diverse} are our lattices? (2) are the candidates encoded in the lattices \textbf{high quality and grammatical}?

\subsection{Datasets \& Base Models}
We obtain all the models and certain baseline decoding methods from the Transformers library \cite{wolf-etal-2020-transformers}. Since our methods are inference techniques without learned components, we do not re-train any models.
For \textbf{summarization}, we use XSum \cite{narayan-etal-2018-dont}, a popular English news summarization dataset. We sample 100 examples from the validation set. The base model we use is \texttt{BART-large-XSum} \cite{lewis-etal-2020-bart}. 
For \textbf{machine translation}, we study our models on the English-French (en-fr) pairs from WMT 2014 \cite{bojar-etal-2014-findings} and Chinese-to-English (zh-en) pair from WMT 2019 \cite{barrault-etal-2019-findings}. We use \texttt{mBART}  \cite{liu-etal-2020-multilingual-denoising}, a state-of-the-art neural machine translation model.
We set the max decoding length to be twice the input length, so it varies per example. 

\subsection{Search Budget}




To fairly compare the resource usage of all methods, we define the search budget as the number of calls to the neural model, equivalent to the number of nodes expanded.\footnote{We incur negligible overhead from rule-based matching in the merging step, as well as the computational costs of computing diversity term in \dbs{} and modifying sampling distributions in sampling methods.}
With beam size \ebs{} and maximum length $T$, beam search methods are given a theoretical budget of \ebs$T$.
We could simply allow best-first search and sampling methods to expand this number of nodes. However, since hypotheses may terminate before they reach EOS, empirically there is a gap between effective length (the average generated hypothesis length) and max length for both beam search and sampling. 
To balance the computation used across the different methods, we apply a correction factor so that the different methods are expanding the same number of nodes in aggregate. We increase the beam size $k$ by 50\% for translation, from 8 to 12, and 25\% for summarization, from 16 to 20, for our baseline methods: \ebs{} to \bs{}, \dbs{}, \topp{}, \temp{}, and \bs\zhang{}. This correction was empirically determined to balance the number of nodes expanded between our method and the baselines. We emphasize that this correction improves the baseline performance relative to our methods.

\subsection{Search Algorithms}
We implemented \greedy{}, \bs{}, \dbs{}, \topp{}, and \temp{} as baseline methods. \topp{0.9} represents nucleus sampling method with $p=0.9$. 
We refer to Appendix~\ref{app:baseline} for detailed descriptions. We also experiment with basic \bfs{} without path recombination, but including our depth-first path completion technique to ensure that finished hypotheses are produced. 
\bs\zhang{} is our implementation of \citet{zhang-etal-2018-exploring}. We integrate \base{} with nucleus sampling and best-first search as \topp{}\base{} and \bfs\base. We also test \bfs{} with the \zip{} strategy. \textleaf\bfs\zip{} is a resource-efficient version of \bfs\zip{} where only 25\% of the search budget is used, exploring what this method can achieve with a lower budget given its more aggressive merges.

\subsection{Evaluation Metrics}
\label{sec-eval-met}

We describe our metrics to evaluate both quality and diversity. Several of our methods build on ROUGE (including R1, R2, RL) \cite{lin-2004-rouge} 
and BLEU \cite{papineni-etal-2002-bleu,post-2018-call} comparing the generated text to references.

\paragraph{Diversity-oriented Metrics}
We evaluate the diversity of generated texts with the following metrics. 
(1) \numpath{} is the average number of unique paths in the produced lattice.\footnote{ Due to the exponentially growing number of paths in some of our models, we cap the number of paths from \nodesos{} to each node to $C=10^4$.} 
(2) Number of unique $n$-grams encoded in the lattice; this captures a different type of diversity than the number of paths, since there could be many paths reusing the same words. 
    \novone{} and \novtwo{} are average number of novel unigrams and bigrams in the graph. 
(3) \selfbleu{} is the average self-BLEU among $m$ samples \cite{zhu2018texygen}. The samples are drawn from a uniform random walk from \nodesos{}. The range of \selfbleu{} is $[0, 100]$.
(4) \edit{} is the average edit-distance among $m$ samples. We set $m=5$ in our experiment.

\begin{table}[t]
\centering
\footnotesize
\setlength{\tabcolsep}{2.4pt}
\begin{tabular}{@{}r|rrrrrccc@{}}
\toprule
                                         & \multicolumn{5}{c}{Diversity} & \taboracle   & \tabsample           & \tabgram \\ 
                & $\uparrow$ & $\uparrow$ & $\uparrow$ & $\downarrow$ & $\uparrow$ & $\uparrow$ & $\geq$         & $\downarrow$    \\
Model           & \numpath   & \novone    & \novtwo    & \selfbleu    & \edit      & R2         & R2             & \errate         \\ \midrule
\greedy                                  & 1      & 22  & 23  & 100 & 0  & 17.3 & 17.3           & 0.5\%    \\
\bs                                      & 20     & 42  & 61  & 87  & 31 & 26.3 & 17.7           & 0.3\%    \\
\dbs                                     & 19     & 59  & 91  & 79  & 53 & 25.5 & 15.9           & 0.5\%    \\
\topp{0.8}                               & 20     & 124 & 237 & 57  & 72 & 30.2 & 14.5           & 0.5\%    \\
\topp{0.9}                               & 20     & 143 & 273 & 50  & 76 & 28.1 & 13.3           & 0.8\%    \\
\temp{1.5}                               & 20     & 170 & 319 & 51  & 82 & 26.6 &  11.6 & 1.4\%    \\
\bfs                                     & 30     & 88  & 167 & 68  & 60 & 30.1 & 15.6           & 0.4\%    \\ \midrule
\multicolumn{9}{c}{\textit{+ Path Recombination}}         \\
\bs \zhang                               & \finecell 4,701  & \badcell 66  & \badcell 118 & \badcell 75  & \badcell 51 & \finecell  33.0 & 16.0           & \goodcell 0.7\%    \\
\topp{0.8}\base{}                        & 52     & \finecell 167 & \finecell 308 & \finecell  53  & \finecell  79 &  28.8 & 13.0           & \finecell 1.0\%    \\
\topp{0.9}\base &  \badcell 36         &  \goodcell 207        &  \goodcell  363        &  \goodcell 50           &  \goodcell  87         & \badcell  25.9       & \badcell  11.0 & \badcell  1.7\% \\
\bfs \base                               & \finecell 7,758  & 111 & 239 & 65  & 64 & \finecell  35.8 & 15.2           & \finecell  0.8\%    \\
\bfs \zip                                & \goodcell  95,744 & \finecell 124 & \finecell  274 & \finecell 53  & \finecell 77 &   \goodcell  36.8 & 13.2           & 1.4\%    \\
\textleaf  \bfs \zip                     & 297    & 58  & 92  & 80  & 49 & 29.2 & 15.2           & 0.8\%    \\ \bottomrule
\end{tabular}
\caption{Results decoding text summaries on XSum. Diversity metrics are rounded to integers to save space. We use $\uparrow$, $\downarrow$ and $\geq$ to denote the desired trend, the higher the better, the lower the better, or good if it passes some threshold. Among the methods with path recombination excluding \textleaf  \bfs \zip, we highlight the \colorbox{RoyalBlue!30}{best}, \colorbox{RoyalBlue!10}{second and third best}, and the \colorbox{BrickRed!15}{worst} one. 
}
\label{tab:xsum-main}
\end{table}


\paragraph{Quality: Grammaticality}
We adopt GECToR 
a neural grammatical error correction model \cite{omelianchuk-etal-2020-gector} to automatically assess the grammaticality of generated texts. 
We report \tabgram \errate (\%), the average number of grammar errors per token, for all English-output experiments.

\paragraph{Quality: Oracle Reference Match}
Given the reference, we find the path with highest ROUGE or BLEU over all found paths. Oracle ROUGE is defined as $\textsc{Or}(\mathcal{Y}, \mathbf{y}^*) = \max_{\mathbf{y} \in \mathcal{Y}}(\text{R2}(\mathbf{y}, \mathbf{y}^*))$.
This metric captures both quality and diversity: the algorithm needs to find something close to the reference, but a diverse lattice will have a higher chance of exhibiting a good candidate all else being equal. 

\begin{table*}[t]
\centering
\footnotesize
\setlength{\tabcolsep}{4pt}

\begin{tabular}{@{}r|rrrrrccc|rrrrrccc@{}} \toprule
            & \multicolumn{8}{c|}{zh-en}                           & \multicolumn{8}{c}{fr-en}                            \\
 &
  \multicolumn{5}{c}{Diversity} &
  \taboracle &
  \tabsample &
  \tabgram &
  \multicolumn{5}{c}{Diversity} &
  \taboracle &
  \tabsample &
  \tabgram \\
                  & $\uparrow$ & $\uparrow$ & $\uparrow$ & $\downarrow$ & $\uparrow$ & $\uparrow$ & $\geq$         & $\downarrow$   & $\uparrow$ & $\uparrow$ & $\uparrow$ & $\downarrow$ & $\uparrow$ & $\uparrow$ & $\geq$         & $\downarrow$    \\

Model &
  \numpath &
  \novone &
  \novtwo &
  \selfbleu &
  \edit &
  \bleu &
  \bleu &
  \errate &
  \numpath &
  \novone &
  \novtwo &
  \selfbleu &
  \edit &
  \bleu &
  \bleu &
  \errate \\ \midrule
\greedy               & 1     & 35  & 40  & 100 & 0   & 24.7 & 24.7 & 0.5\% & 1       & 28  & 31  & 100 & 0  & 40.0 & 40.0 & 0.9\% \\
\bs                  & 12    & 45  & 63  & 95  & 20  & 32.2 & 25.0 & 0.2\%  & 12      & 37  & 50  & 93  & 13 & 52.6 & 38.1 & 1.0\%  \\
\dbs                 & 11    & 55  & 84  & 89  & 59  & 29.7 & 20.5 & 0.5\%  & 11      & 45  & 67  & 88  & 37 & 46.4 & 30.5 & 1.1\%  \\
\topp{0.8}           & 12    & 94  & 188 & 72  & 82  & 31.5 & 17.5 & 0.7\%  & 11      & 62  & 107 & 80  & 46 & 51.0 & 31.2 & 1.0\%  \\
\topp{0.9}           & 12    & 110 & 226 & 67  & 94  & 30.4 &  15.8 & 0.9\%  & 12      & 75  & 134 & 77  & 57 & 48.3 &  27.4 & 1.2\%  \\
\temp{1.5}           & 12    & 140 & 280 & 62  & 105 & 27.0 &  12.7 & 1.3\%  & 12      & 102 & 184 & 69  & 71 & 43.7 &  21.6 & 1.6\%  \\
\bfs                  & 18    & 60  & 104 & 86  & 54  & 32.7 & 20.7 & 0.5\% & 27      & 59  & 102 & 84  & 37 & 53.2 & 33.7 & 1.1\% \\ \midrule
\multicolumn{17}{c}{\textit{+ Path Recombination}}            \\
\bs \zhang            & \finecell  18,336 &\badcell  64  & \badcell  117 & \badcell  77  & \badcell  65  &  \finecell 40.1 & 19.1 & \goodcell 0.8\%  & \finecell 16,729  & \badcell 59  & \badcell 107 & \badcell 77  & \badcell 43 & \finecell 61.2 & \goodcell 28.2 & \finecell 1.3\%  \\
 \topp{0.8}\base        & 81    &  \finecell 138 &  \finecell 263 &  \finecell 67  &  \finecell 91  & 26.8 &  13.9 &  \finecell 1.1\% & 344     & \finecell 140 & \finecell 246 & \finecell 64 & \finecell 67 & 48.2 & \finecell 26.6 & \goodcell 1.2\% \\
  \topp{0.9}\base       & \badcell 38    &  \goodcell 188 & \goodcell  343 &  \goodcell  58  & \goodcell  114 & \badcell 23.9 & \badcell 10.6 & \badcell  1.7\% & \badcell 123     &  \goodcell 205 &  \goodcell  352 &  \goodcell  55  &  \goodcell  92 & \badcell 41.1 &  20.2 & 2.1\% \\
\bfs \base            &  \finecell 17,535 & 81  & 171 & 76  & 72  &  \finecell 42.1 & 19.4 &  \finecell 0.9\% & \finecell 47,577  & 85  & 193 & 68  & 52 & \goodcell 64.6 & \finecell 25.3 & \finecell 1.6\% \\
\bfs \zip             & \goodcell  59,020 &  \finecell 94  &  \finecell  205 &  \finecell 66  & \finecell  88  & \goodcell 42.4 &  15.5 & 1.4\% & \goodcell 146,163 & \finecell 111 & \finecell 259 & \finecell 56  & \finecell 63 & \finecell 56.8 & \badcell  16.9 & \badcell 2.5\% \\
\textleaf  \bfs \zip  & 511   & 50  & 75  & 89  & 38  & 33.0 & 21.2 & 0.7\% & 4,531   & 50  & 81  & 82  & 35 & 59.5 & 29.4 & 1.4\% \\ \bottomrule
\end{tabular}
\caption{Results on WMT14 Fr-En and WMT19 Zh-En.  Columns are the same as for summarization, although BLEU is used instead of ROUGE. Trends are roughly similar, with \bfs\base{} providing high diversity at good quality and \textleaf \bfs\zip{} offering a strong tradeoff between computational resources and diversity.}
\label{tab:x-en}
\end{table*}

\paragraph{Quality: Average Reference Match}

Although our method focuses on deriving diverse text summaries or translations, we aim to guarantee that the generated text is highly relevant to the generation target and is of high quality in general. We sample 1,000 paths from the lattice with replacement and evaluate the average ROUGE or BLEU compared to the reference. 
We denote this metric as \tabsample.


\section{Results}
\label{sec:results}

\paragraph{Text Summarization}
We present the experimental results on the dev set of XSum in Table~\ref{tab:xsum-main}. Full results are kept in Table~\ref{tab:xsum-full} for reference. 
Among non-recombination methods, \bs{} and \dbs{} are the least diverse methods. Sampling based methods including \temp{} are generally more diverse, but the oracle ROUGE is lower than that of \bfs{}. Given the sacrificed text quality (lower sample ROUGE and more grammar errors) of sampling based methods, we argue that \textbf{modified best-first search is a strong decoding strategy even without path recombination}. 
The bottom half shows all methods with path recombination techniques. \textbf{Recombination significantly improves the diversity of generated outputs}, with a much higher number of paths. The \selfbleu{} of the recombination variants are lower than their non-recombination counterparts.

In terms of search quality, the proposed \bfs\base{} and \bfs\zip{} methods obtain significantly higher oracle ROUGE compared to all other methods. We show these results later in Figure~\ref{fig:large-beam-oracle}: our approach can find much better oracle solutions, even compared with beam search method with quadruple the amount of computational resources.
The design of the oracle ROUGE metric is also motivated by a real use case: if you want a specific summary (e.g., a summary covering a specific entity or topic), does it exist in the search graph? 
Higher oracle ROUGE indicates a closer match, meaning a strategy using some kind of reranking model could help find the user's preferred outputs.


\paragraph{Comparison: \base{} \& \zip}
The \zip{} method yields even more diverse output at the cost of text quality. There are a few reasons for this: 1) recombination of more nodes makes the lattice denser, increasing the number of paths but also potential errors; 2) elimination of unexplored children from merged branch reduces the waste of exploration which means \zip{} can explore more hypotheses than \base{}. With the same amount of computational resources, \zip{} explores a larger search space while \base{} explores a smaller collection more reliably. 
\textleaf{}\zip{} exploits the efficiency of \zip{} to achieve high diversity, and by searching through fewer states, it manages to achieve higher quality as well.

\paragraph{Machine Translation}
We show the result on machine translation in Table~\ref{tab:x-en} and \ref{tab:en-fr}. 
Results on translation tasks show the consistent gains of diversity from path recombination models. 
In Table~\ref{tab:x-en}, we show two translation tasks where the target language is English. \bfs\base{} works better than \bfs\zip{} because it disables some aggressive and bad merges which explores bad hypotheses. Compared to summarization, we found the search space in MT to be more constrained, so there was less room for aggressive merging and exploration to improve over \base{}.
Our lower-resource method, \textleaf\bfs\zip{} approach, actually performs quite well on most metrics with only 25\% of search budget. It has better diversity performance than any non-recombination methods, and comes with quality better than most of the recombination methods. The usage of \bfs{} and path recombination methods like \bfs{}\base{} and \bfs\zip{} is promising for being able to find a better cost-diversity tradeoff in MT.

\begin{figure}[t]
    \centering
    \footnotesize
    \includegraphics[width=0.42\textwidth]{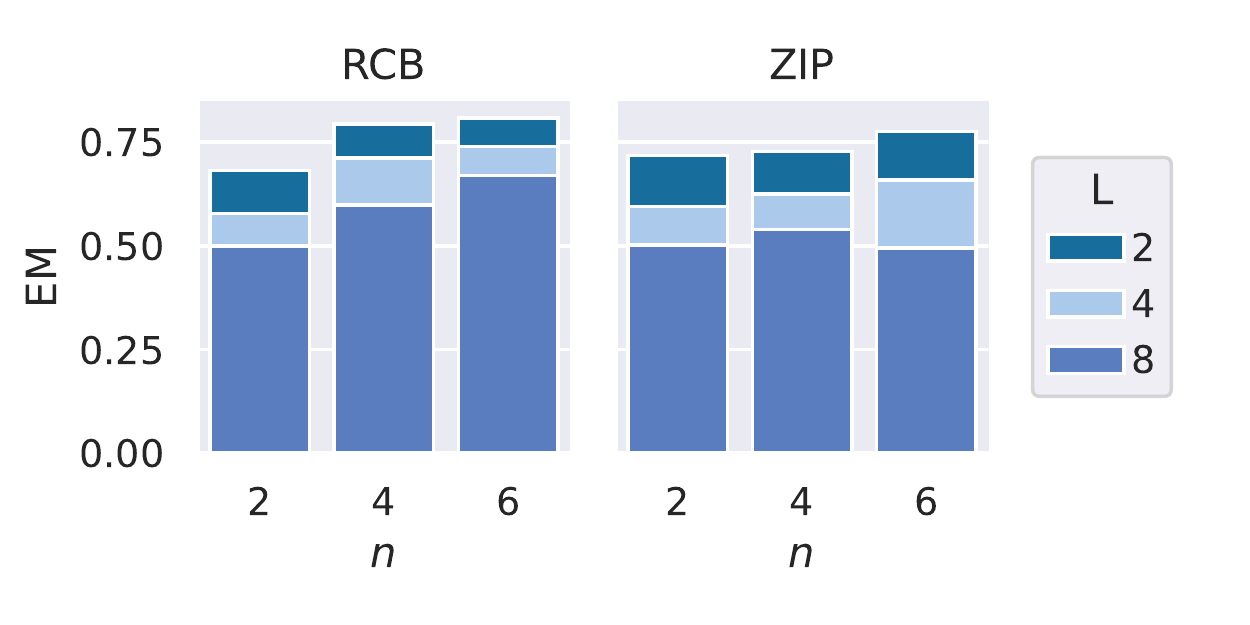}
    \caption{Empirical verification of merging criteria. We experiment with $n=\{2,4,6\}$ for $n$-gram suffix matching. We sample 1,000 recombinations from \bfs\base{} and \bfs\zip{} respectively, and run greedy inference based on merged hypotheses. We use Exact Match (EM) to measure how often two merged hypotheses give the same greedy future generations considering the next $L$ tokens after the merge.}
    \label{fig:check}
\end{figure}

\begin{figure}[t]
    \centering
    \includegraphics[width=0.44\textwidth]{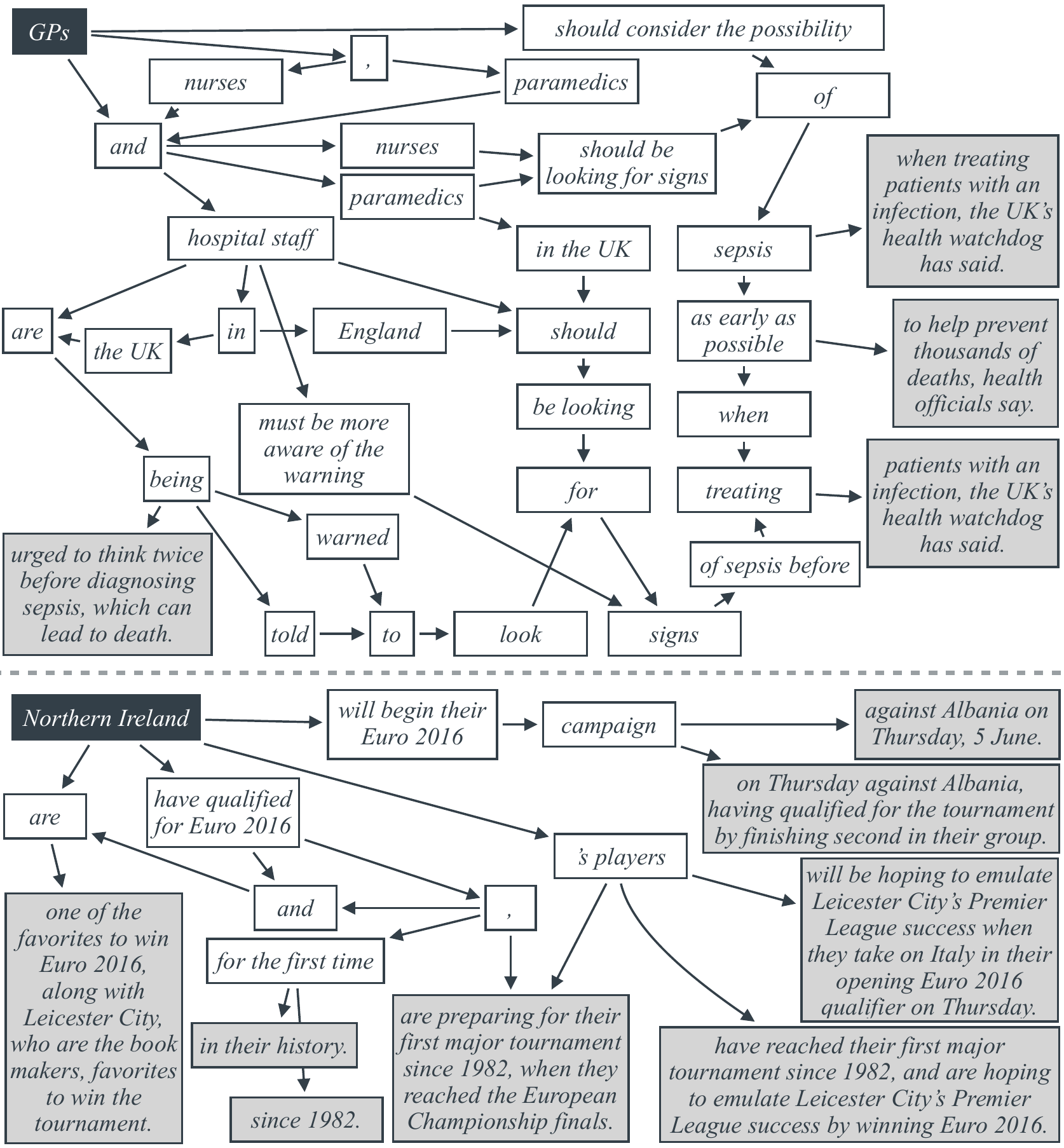}
    \caption{Two examples on XSum by \textleaf\bfs\zip{}. The start of sentence is denoted in dark color, and all the endings are in gray. We combine tokens to phrases when possible for visualization purpose. More examples are presented in Appendix.~\ref{app:example}. }
    \label{fig:error}
\end{figure}

\paragraph{Validating the Merging Criterion}

Our merging criterion is fundamentally an approximation of the equivalence criteria described in Section~\ref{sec_rec}. 
Our question is: \textbf{what fraction of nodes merged by our merging criterion satisfy the weak equivalence assumption?}
We conduct an experiment to verify this on XSum. 
We compute the greedy completion for $L$ timesteps and check whether the continuation of the base candidates would be the same.
In Figure~\ref{fig:check}, we show the fraction of merged pairs for which the generations match exactly under three values of the recombination criterion. For \bfs\base{}, when using $n=4$ the greedy continuation over 4 timesteps is the same 71.2\% of the time. For \bfs\zip{} it is the same 62.5\% of the time. 
Following the weak equivalence criterion is a strong indication that these hypotheses can admit many of the same continuations. \base{} is more reliable than \zip{}, but both methods show moderate adherence to the equivalence criterion.

\paragraph{Error Analysis \& Visualization}
In Figure~\ref{fig:error}, we present two examples on XSum by \textleaf\bfs\zip{}. The upper example has more word level recombination and paraphrasing while the bottom one has more ways of ending and more diverse content coverage. 
We show more examples on both summarization and translation in Appendix.~\ref{app:example}.

We manually examine the output and found a few common types of errors introduced by our algorithm.
(1) Factual errors at high entropy nodes. Our approach assumes that high-scoring candidates under the model are good quality, but this assumption is violated in certain cases, like when the model attempts to hallucinate information. For example, the prefix ``\emph{The company, founded in}'' will cause the model to guess answers like ``\emph{1989}'' or ``\emph{1999}''. Encoding all of these in the lattice is incorrect. However, we did not see significant factual errors introduced by merging specifically. 
(2) Aggressive bad merges. In the upper example in Figure~\ref{fig:error}, the cluster of ``\emph{GPs}'', ``\emph{nurses}'', ``\emph{paramedics}'' is an example case. The lattice encodes paths like ``\emph{GPs, nurses and nurses should ...}''. This could be fixed by heuristics or rules in future work.


\section{Related Work}

The techniques used in this work partially reflect an outgrowth of a few lines of literature: understanding the behavior of text generation models \cite{xu-etal-2020-understanding-neural,xu-durrett-2021-dissecting,zhong-etal-2021-adapting-language}, investigations into beam search \cite{stahlberg-byrne-2019-nmt,meister-etal-2020-beam}, and studies of diversity in generation.

In terms of search strategies, best-first beam search \cite{meister-etal-2020-best} is a method integrating best-first search with beam search.
Some other variants of search have also been studied in previous work \cite{meister-etal-2021-determinantal,meister-etal-2021-conditional}. 
Beam search has been critically examined in some recent work \cite{huang-etal-2017-finish,stahlberg-byrne-2019-nmt}, but largely of focused on specific challenges in MT.


As for diverse generation, neural text degeneration has been discussed in \citet{radford2019language,Holtzman2020The,Welleck2020Neural}, which led to an interest in diverse generation models.
Diverse text generation has been studied in previous work \cite{yu2017seqgan}, including in dialogue \cite{li-etal-2016-diversity}, story generation \cite{fan-etal-2019-strategies}, and particularly paraphrasing \cite{iyyer-etal-2018-adversarial,goyal-durrett-2020-neural}. Our method can also diversify content coverage \cite{gehrmann-etal-2018-bottom} and word choice \cite{cao-wang-2021-inference}.

\section{Discussion \& Conclusion}

We presented an algorithm for decoding in text generation with two main components: best-first search to more efficiently structure exploration of the space and hypothesis recombination to encode summaries in a lattice structure. We showed that across summarization and machine translation, these lattices successfully encode large numbers of high-quality generation options.

There are a few limitations of our method. 
First, we currently benchmark these techniques using number of nodes expanded, not wall clock time. There are strategies for parallelizing the \bfs{} expansion \cite{shu-nakayama-2018-improving}, but it remains to be seen how this parallelism compares to the parallelism achieved by beam search. Regardless, the dramatically larger number of hypotheses we return outweighs efficiency differences for now.
Second, we focus on auto-regressive methods in this paper. However, we believe our framework could also be applied and adopted to non auto-regressive generation models \cite{song-etal-2021-new}.

\section*{Acknowledgments}
We would like to thank Shuyang Cao, Eunsol Choi, Tanya Goyal, Jonathan Kummerfeld, Jessy Li, Yasumasa Onoe, Xi Ye, and Zhisong Zhang for input and feedback on this work. 
This work was principally supported by a gift from Amazon, as well as NSF Grant IIS-1814522 and a gift from Salesforce Inc. Thanks to the anonymous reviewers for their helpful feedback.

\bibliography{anthology2021,custom}
\bibliographystyle{acl_natbib}

\appendix

\clearpage

\section{Inadequacies of Beam Search: Poor Scaling Behavior}
\label{app-beam-fail}

\begin{figure}[t]
    \centering
    \footnotesize
    \includegraphics[width=0.4825\textwidth]{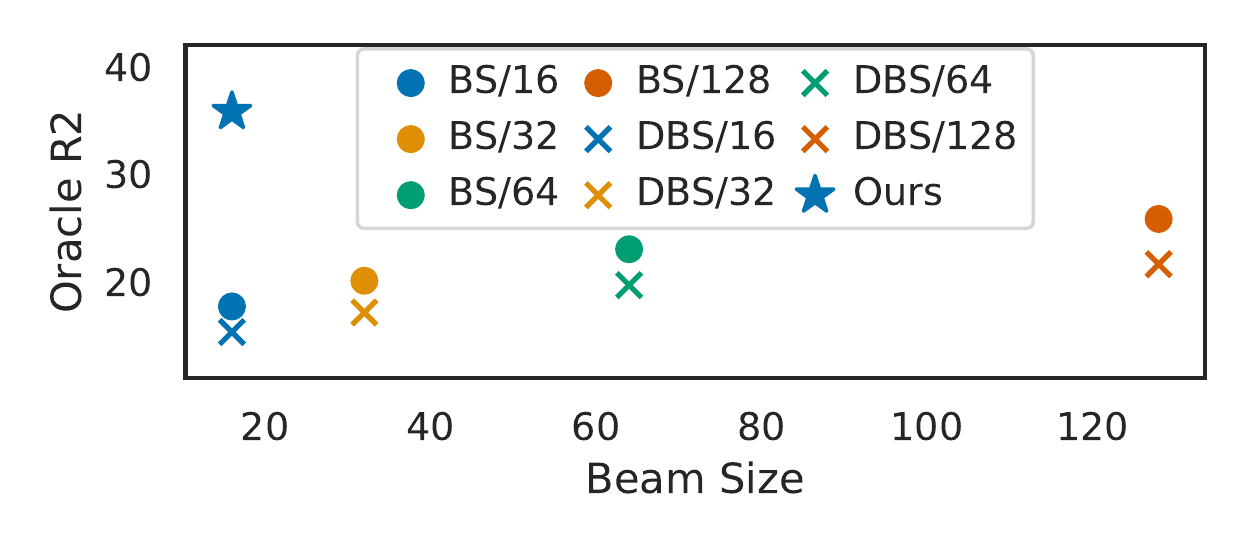}
    \caption{Oracle R2 of \bs{}/\dbs{} with larger beam size $k$. Blue star represents \bfs\base{} with equivalent $k=16$. }
    \label{fig:large-beam-oracle}
\end{figure}

In spite of the issues with beam search that we describe in the main text, perhaps beam search could still be viable with larger beam sizes if more computational resources are available.
We experiment with beam sizes of $2^{ \{4,5,6,7 \}}$ and see how diversity scales with beam size. 
In Figure~\ref{fig:large-beam-tradeoff}, we found that an exponential increase of beam size does not lead to a strong increase of number of novel bigram in beam search. In \dbs{}, the diversity does ramp up, but the quality of the generated text decreases substantially.  \textbf{For \bs{} and \dbs{}, increasing beam size is not an effective solution for better diversity.} 
We compare the oracle R2 of \bs{}/\dbs{} with larger beam size in Figure~\ref{fig:large-beam-oracle}. The oracle R2 increases slowly as $k$ doubles, but our model  \bfs\base{} with $k=16$ achieves 35.8, much higher than all \bs{}/\dbs{} cases.


\label{app:bs-scale}


\section{Strong Equivalence of Path recombination}
\label{app:strong}
In the strictest form, recombining two hypotheses assumes the following equivalence between them:
\begin{define}[Strong equivalence]
\label{assump-strong}

Let $\mathbf{a}$ and $\mathbf{b}$ be two prefix strings starting with \nodesos{}. $\mathbf{a}$ and $\mathbf{b}$ are strongly equivalent if $P(\mathbf{y} \mid \mathbf{a}) = P(\mathbf{y} \mid \mathbf{b})$ holds for all $\mathbf{y}$.
\end{define}
\noindent
Merging such states in the search tree is valid with no loss of information, as any expanded node will receive the same score under both prefixes.
However, this assumption is not realistic since seq2seq models condition on the entire sequence so far, and any tiny perturbation changes the predicted distribution. To relax the assumption, we then propose the weak alternative.

\section{Proof of Theorem~\ref{theorem:can}}
\label{app:proof}
Proof by induction. Base case: we begin with just \nodesos{} in the lattice, which has exactly one canonical path consisting of itself.

Inductive case: assume every node in the lattice has exactly one canonical path. We have to consider two cases when expanding a node in the lattice:

(1) Expanding the node $n$ as normal. In this case, $n$ is on the search frontier due to its parent node $n'$ being expanded, which establishes a \gen{} edge from $n'$ to $n$. Since $n'$ has exactly one canonical path, $n$ then has exactly one canonical path consisting of the canonical path to $n'$ extended to $n$.

(2) Applying recombination. This operation only adds \mrg{} edges and deletes nodes, neither of which have any impact on the canonical paths.

\section{Implementation Details: \zip{}}
\label{app:zip}
We summarize the key differences of \zhang{}, \base{} and \zip{} in Table~\ref{tab:keydiff}. 
In \zip{}, nodes that are already expanded might be removed from the lattice due to recombination. For example, in Figure~\ref{fig:fullrecomb}, node 6 and 7 are removed in this fashion. In general, we handle this by re-mapping the eliminated node to its surviving counterpart. Any reference to node 7 is routed to node 3, or whatever node 3 is mapped to. This procedure is defined and implemented as a union-find data structure. 

\begin{figure}[t]
    \centering
    \footnotesize
    \includegraphics[width=0.4825\textwidth]{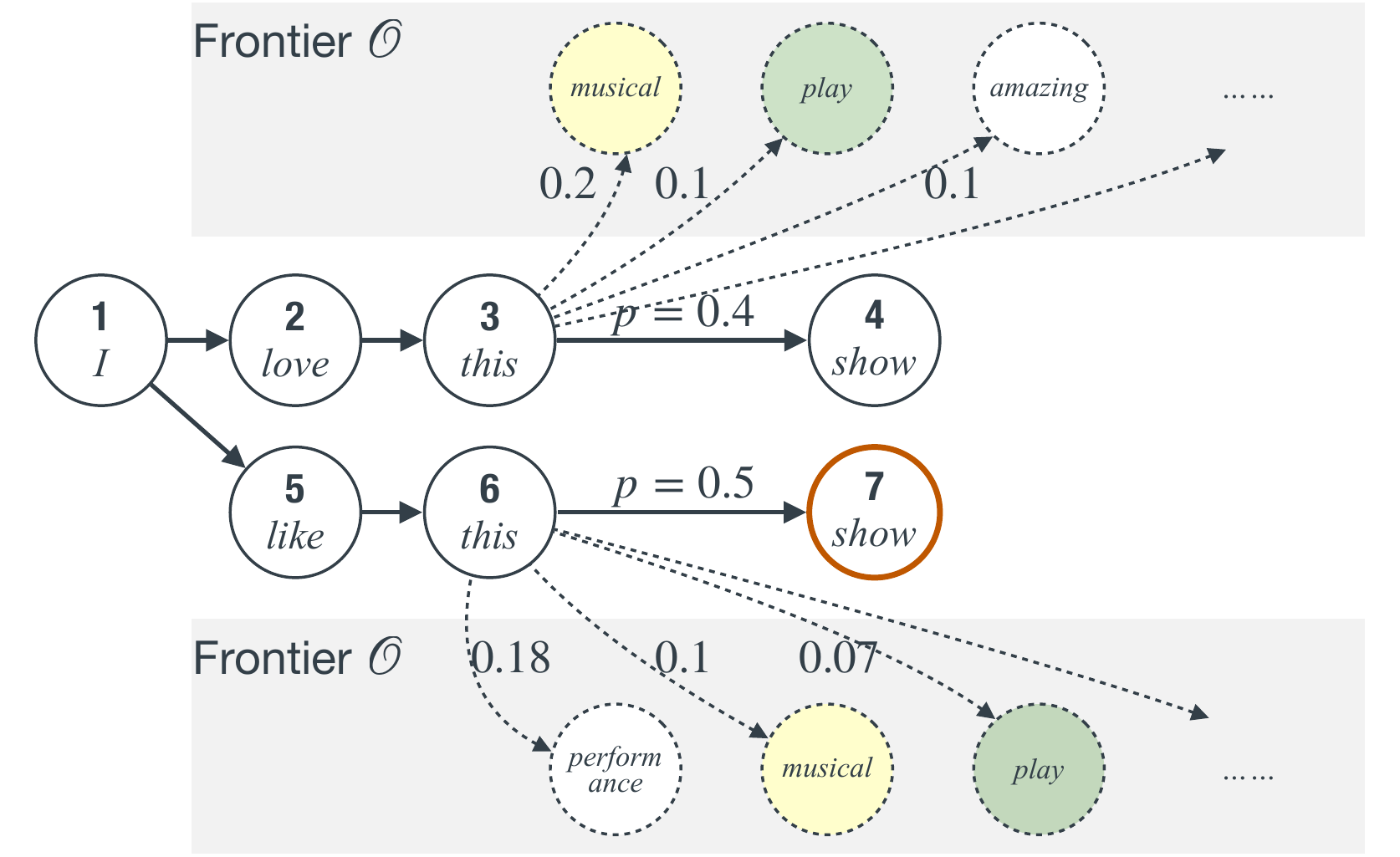}
    \caption{An illustration of removing unexplored hypotheses from search frontier in \zip{}.}
    \label{fig:unexp}
\end{figure}

\paragraph{Deduplication of Unexplored Successors}
After the \zip{} procedure, we also remove the unexplored successors of the merged nodes, like node 6, 7, and 8 in Fig.~\ref{fig:fullrecomb}. 
We show a more detailed example in Figure~\ref{fig:unexp}. 
In \zip{}, we will merge node 3 and node 6. If we take a closer look at the successors of these two nodes, the distributions could be similar and in fact are expected to be if the equivalence criteria hold. We remove the unexplored direct successors of the merged node as part of the merging process, and the surviving node (node 3) captures these with similar probabilities regardless. 

\section{Baselines}
\label{app:baseline}
\paragraph{\greedy} is the deterministic greedy decoding method that always selects the highest probability token as prediction. The equivalent beam size for this approach is 1 since we only run one pass.
\paragraph{\bs{} \& \dbs} stand for beam search and its variant diverse beam search \cite{vijayakumar2016diverse}. 
In our configuration, we use Hamming distance as the diversity function and set the diversity strength to 1.5, following  \citet{vijayakumar2016diverse}.

\paragraph{\topp{}} is the nucleus sampling method proposed in \citet{Holtzman2020The}, which encourages quality by truncating the distribution over the vocabulary with a parameter $p$ before sampling. We experiment it with $p=0.9$ and $p=0.8$.

\paragraph{\temp{}} changes the temperature of softmax function to reshape the prediction distribution \cite{ficler-goldberg-2017-controlling}.  We set the temperature parameter $\tau =1.5$ so the prediction picks more low-scored tokens than $\tau = 1$.

\section{Implementation Details: Beam Search}
\label{app:beam}
In our beam search implementation, the size of the search frontier $\mathcal{O}$ is up to the beam size $k$. 
When a path is completed, we move it from the search frontier $\mathcal{O}$ to a completed set $\mathcal{F}$ to free up the beam for exploring unfinished hypotheses. Naturally, finished hypotheses $\mathcal{F}$ in the end can be of variable length.
After reaching the max generation step $T$, we sort all hypotheses in $\mathcal{F}$ according to the model score. Following common practice in libraries such as Transformers \cite{wolf-etal-2020-transformers}, we return a number of completed hypotheses equal to the beam size.

\begin{table*}[t]
\centering
\footnotesize
\begin{tabular}{@{}r|rrrrr|ccccccc@{}}
\toprule
                                         & \multicolumn{5}{c|}{Diversity} & \multicolumn{3}{c}{Oracle} & \multicolumn{3}{c}{Sample} & \tabgram \\ 
 &
  $\uparrow$ &
  $\uparrow$ &
  $\uparrow$ &
  $\downarrow$ &
  $\uparrow$ &
  $\uparrow$ &
  $\uparrow$ &
  $\uparrow$ &
  $\geq$ &
  $\geq$ &
  $\geq$ &
  $\downarrow$ \\
Model &
  \numpath &
  \novone &
  \novtwo &
  \selfbleu &
  \edit &
  R1 &
  R2 &
  RL &
  R1 &
  R2 &
  RL &
  \errate \\ \midrule
\greedy                   & 1      & 22  & 23  & 100 & 0  & 41.4    & 17.3    & 33.5   & 41.4    & 17.3    & 33.5   & 0.5\%                   \\
\bs                       & 20     & 42  & 61  & 87  & 31 & 47.6    & 26.3    & 40.3   & 41.5    & 17.7    & 33.6   & 0.3\%                   \\
\dbs                      & 19     & 59  & 91  & 79  & 53 & 47.0    & 25.5    & 39.1   & 38.5    & 15.9    & 30.3   & 0.5\%                   \\
\topp{0.8}              & 20     & 124 & 237 & 57  & 72 & 50.4    & 30.2    & 44.2   & 37.4    & 14.5    & 29.5   & 0.5\%                   \\
\topp{0.9}              & 20     & 143 & 273 & 50  & 76 & 48.0    & 28.1    & 42.2   & 36.1    & 13.3    & 28.5   & 0.8\%                   \\
\temp{1.5}              & 20     & 170 & 319 & 51  & 82 & 45.0    & 26.6    & 38.5   & 34.1    &  11.6    & 26.3   & 1.4\%                   \\
\bfs                      & 30     & 88  & 167 & 68  & 60 & 50.8    & 30.1    & 44.0   & 39.0    & 15.6    & 30.8   & 0.4\%                   \\ \midrule
\multicolumn{13}{c}{\textit{+ Path Recombination}}                                                                                                                    \\
\bs \zhang & 4,701   & 66  & 118 & 75  & 51 & 52.2    & 33.0    & 45.7   & 40.0    & 16.0    & 32.3   & 0.7\%                   \\
\topp{0.8}\base{} &
  52 &
  167 &
  308 &
  53 &
  79 &
  49.0 &
  28.8 &
  41.8 &
  35.0 &
  13.0 &
  27.8 &
  1.0\% \\
\topp{0.9}\base &
  36 &
  207 &
  363 &
  50 &
  87 &
  44.6 &
  25.9 &
  38.7 &
  32.1 &
   11.0 &
  25.1 &
   1.7\% \\
\bfs \base & 7,758  & 111 & 239 & 65  & 64 & 55.2    & 35.8    & 49.3   & 38.5    & 15.2    & 30.8   & 0.8\%                   \\
\bfs \zip  & 95,744 & 124 & 274 & 53  & 77 & 55.6    & 36.8    & 48.8   & 36.8    & 13.2    & 28.7   & 1.4\%                   \\
\textleaf  \bfs \zip &
  297 &
  58 &
  92 &
  80 &
  49 &
  49.6 &
  29.2 &
  42.8 &
  38.8 &
  15.2 &
  31.0 &
  0.8\% \\ \bottomrule
\end{tabular}
\caption{Full results for all methods decoding text summaries on XSum. }
\label{tab:xsum-full}
\end{table*}

\begin{table}[t]
\centering
\footnotesize
\begin{tabular}{@{}c|cccc@{}}\toprule
 Method  & \textsc{Algos}         & \textsc{Cand}           & \textsc{Len} & \textsc{Dedup} \\ \midrule
\bs\zhang  & \bs       & last step & 1                    & N                           \\ 
\base & any      & all      & 1                    & N                           \\ 
\zip & any     & all      & $n$                   & Y                          \\ \bottomrule
\end{tabular}
\caption{Key differences in path recombination methods. BS\zhang{} is the recombination method used in \citet{zhang-etal-2018-exploring}. \textsc{Algos} shows which search or decoding methods this method is used with. \textsc{Cand} is where the merge candidates come from in the lattice. \textsc{Len} reflects how many nodes are recombined per operation.  \textsc{Dedup} denotes whether duplicates on the merged branch will be removed from heap.}
\label{tab:keydiff}
\end{table}

\section{Results of WMT14 English to French}

Table~\ref{tab:en-fr} shows an additional experiment on English-French. We do not evaluate on grammaticality due to the GECToR model being specialized to English. The results show broadly similar trends as those in Table~\ref{tab:x-en}, discussed in the main text.

\begin{table}[t]
\centering
\setlength{\tabcolsep}{4pt}
\footnotesize
\begin{tabular}{@{}r|rrrrrcc@{}}
\toprule  & \multicolumn{5}{c}{Diversity}       & \taboracle &
 \tabsample \\
   & $\uparrow$ & $\uparrow$ & $\uparrow$ & $\downarrow$ & $\uparrow$ & $\uparrow$ & $\geq$  
  \\ 
Model &
\numpath  &
\novone &
 \novtwo &
\selfbleu &
  \edit &
 \bleu &
 \bleu \\ \midrule
\greedy              & 1      & 32  & 35  & 100 & 0   & 28.5                       & 28.5                       \\
\bs                  & 12       & 42      & 57      & 93        & 13    & 37.8                     & 27.5          \\
\dbs                 & 10       & 51      & 73      & 89        & 38    & 33.1                     & 22.7          \\
\topp{0.8}           & 12       & 95      & 171     & 72        & 56    & 35.4                     & 20.4          \\
\topp{0.9}           & 12       & 116     & 214     & 66        & 73    & 33.4                     &  17.6          \\
\temp{1.5}           & 12       & 150     & 274     & 61        & 89    & 28.4                     &  13.1          \\
\bfs   & 17     & 62  & 98  & 85  & 35  & 38.8                       & 25.0                       \\  \midrule
 \multicolumn{8}{c}{\textit{+ Path Recombination}}   \\
\bs\zhang            & \finecell 17,508   & \badcell 67      & \badcell 117     & \badcell 78        & \badcell 40    & \finecell 46.4  & \goodcell 21.2          \\
 \topp{0.8}\base{}    & 59     & \finecell 151 & \finecell 261 & \finecell 67  & \finecell 78  & 29.3                       &  16.3                      \\
  \topp{0.9}\base{}        & \badcell 32     & \goodcell 190 & \goodcell 317 & \goodcell 53  & \goodcell 101 & \badcell 26.9                       & \badcell 12.6                       \\
\bfs\base     & \finecell 18,663 & 90  & 180 & 74  & 42  & \goodcell 46.6  & \finecell 20.8 \\
 \bfs \zip           & \goodcell 49,507 & \finecell 104    & \finecell 213    &\finecell 65     & \finecell 53 & \finecell 45.9 & \finecell 16.7   \\
\textleaf  \bfs \zip   & 386    & 49  & 70  & 88  & 25  & 39.5                       & 25.7                       \\ \bottomrule
\end{tabular}
\caption{Results on machine translation WMT14 English to French. \bfs\base{} and \bfs\zip{} are strong in both diversity and quality. }
\label{tab:en-fr}
\end{table}

\section{Examples}
\label{app:example}
\begin{figure*}[]
    \centering
    \footnotesize
    \includegraphics[width=\textwidth]{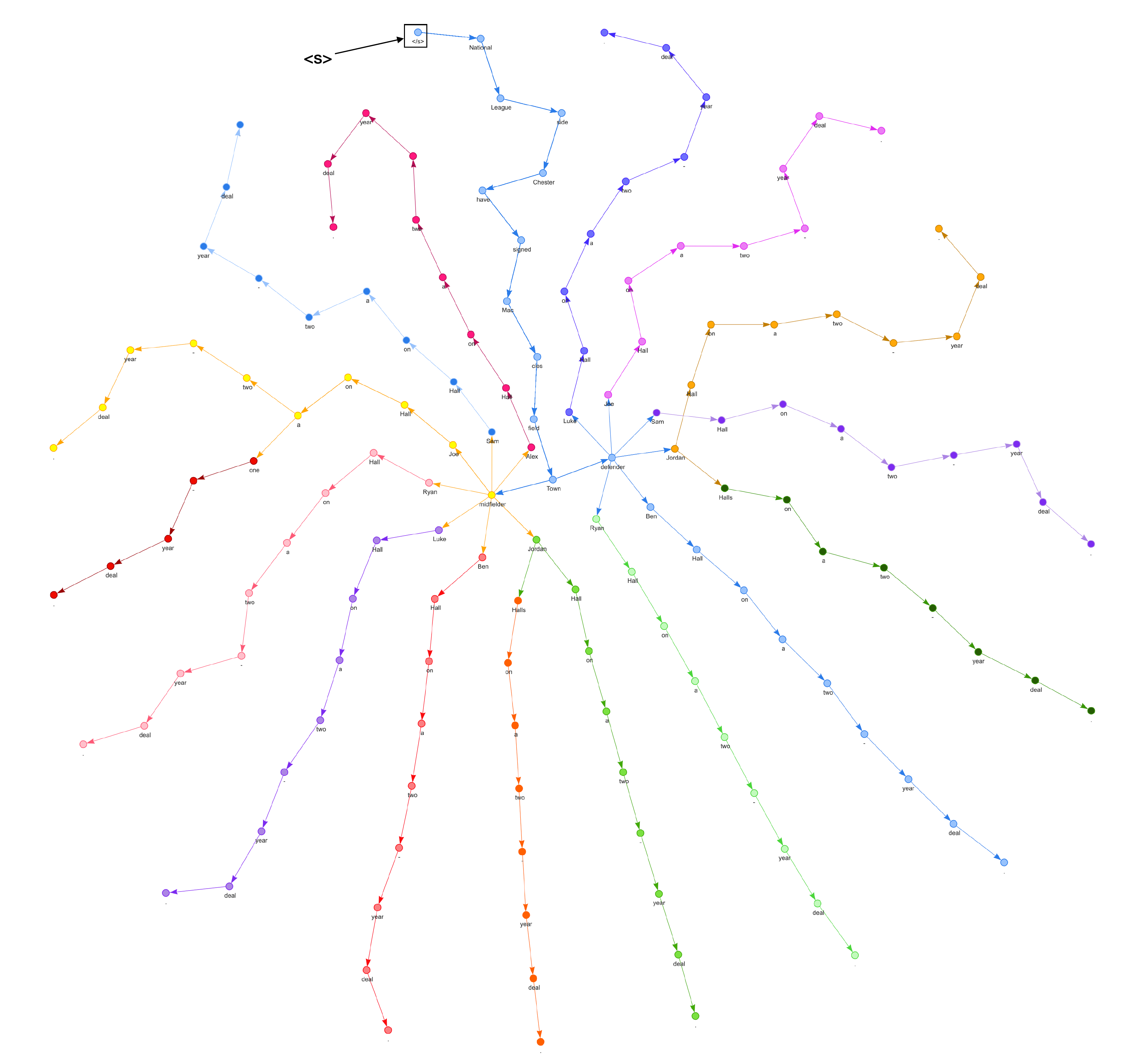}
    \caption{Visualization of one example output for beam search on XSum. \nodesos{} is labeled. Each color represents one unique ending.  }
    \label{ex-bs}
\end{figure*}
\begin{figure*}[]
    \centering
    \footnotesize
    \includegraphics[width=\textwidth]{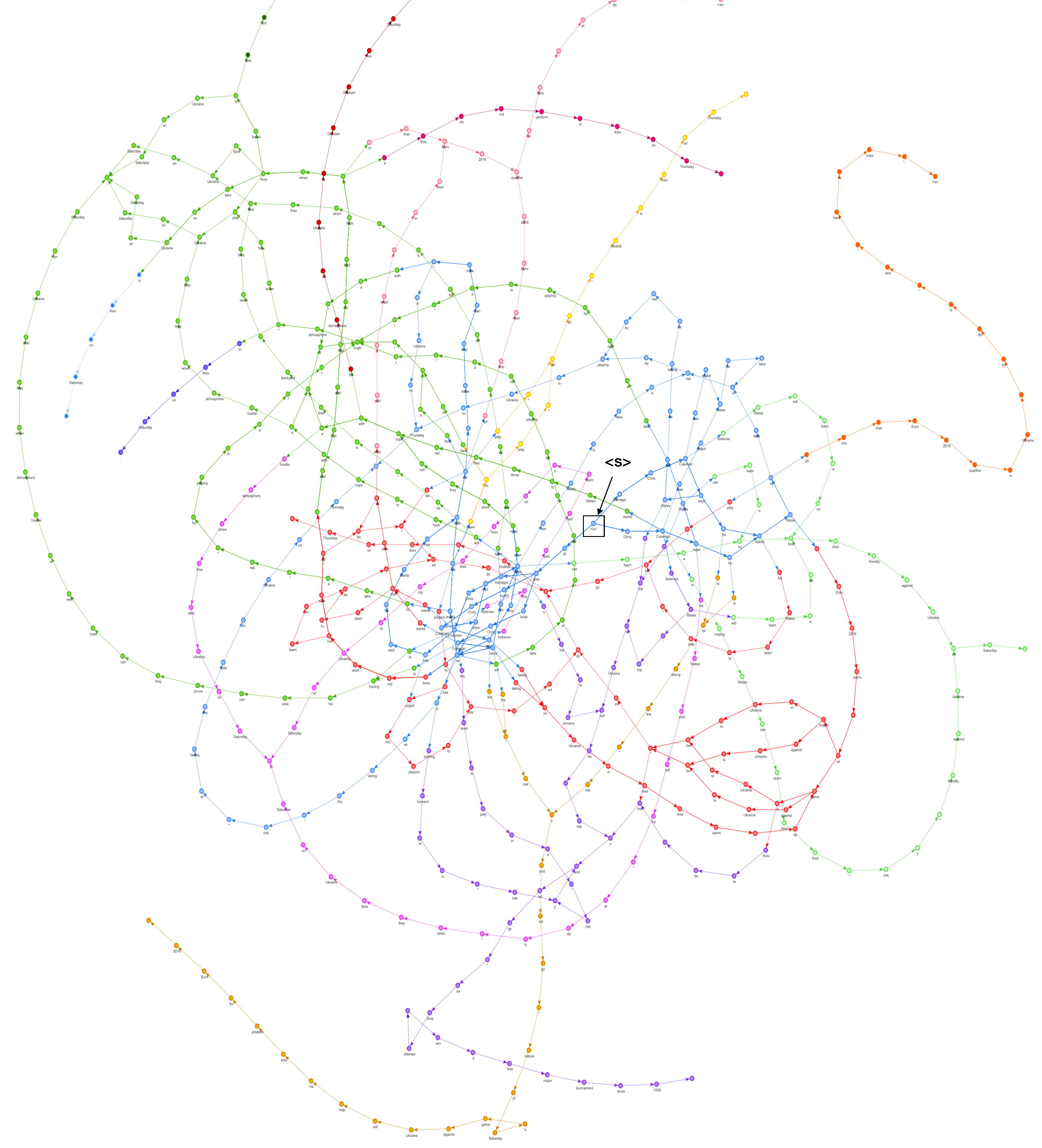}
    \caption{Visualization of one example output for \bfs\base{} on XSum. }
    \label{ex-base}
\end{figure*}

\begin{figure*}[]
    \centering
    \footnotesize
    \includegraphics[width=\textwidth]{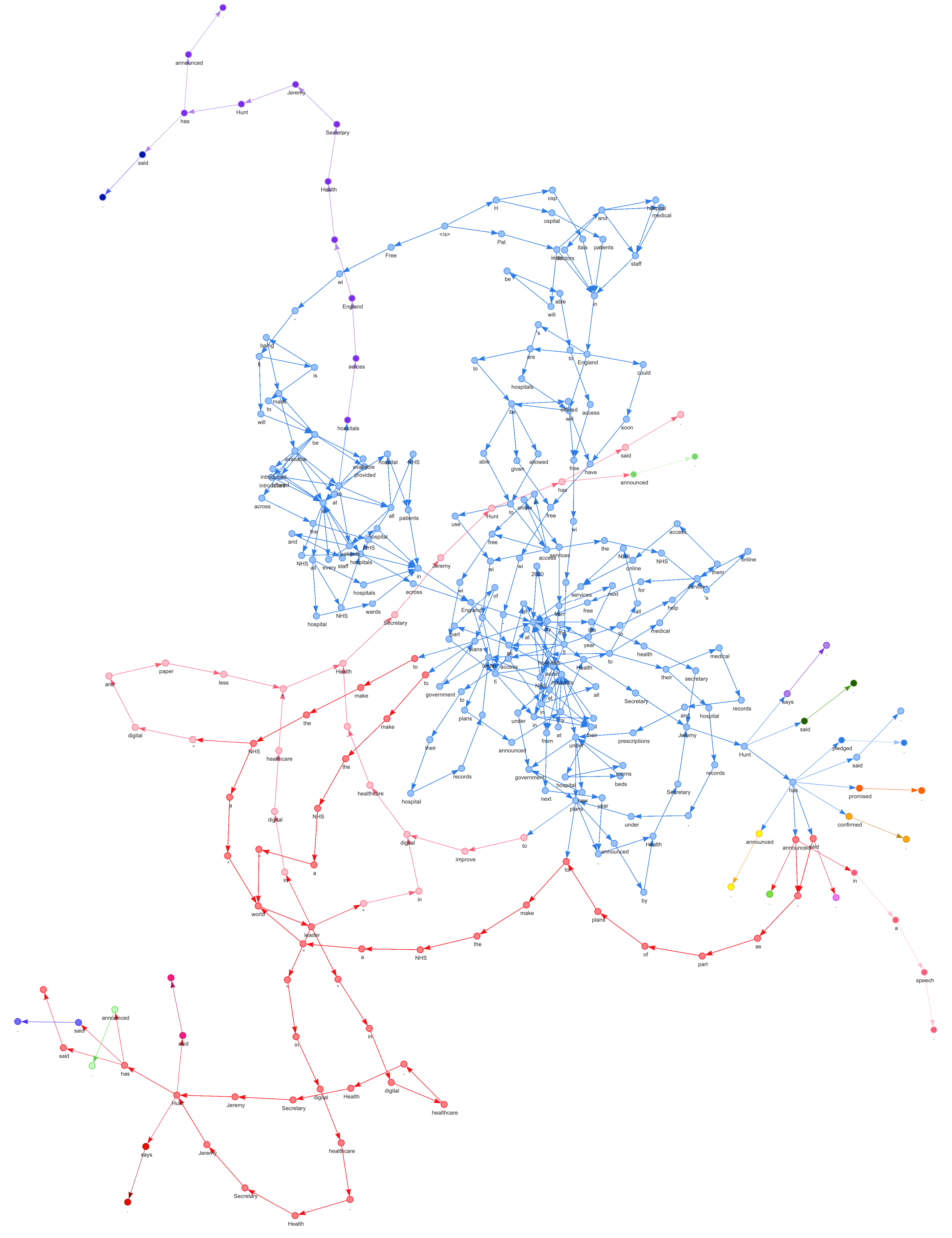}
    \caption{Visualization of one example output for \bfs\zip{} on XSum. }
    \label{ex-zip}
\end{figure*}
We show three examples with visualization in Figure~\ref{ex-bs},\ref{ex-base} and \ref{ex-zip}. We use \texttt{PyVis} as the visualization tool.\footnote{\url{https://github.com/WestHealth/pyvis}}
More examples are available at \url{anonymized}. 

\section{Computational Considerations}

\paragraph{Resources Used} All experiments were conducted on a server with 32GB RAM and Intel Xeon E5-2630 CPU, using a single NVIDIA GTX1080. The total computational budget in GPU hours is within 50 hours for experiments in text summarization and machine translation.

\paragraph{Memory and Runtime} Although the final lattices returned encode large numbers of paths, they do not take large amounts of memory. Because the number of nodes in a lattice is no larger than the number of node expansion operations during beam search, it is always less than the search budget and can be stored compactly.

Moreover, the wall clock time of our BFS-Recomb strategy is manageable, on the order of between 1 and 10 seconds for summarization. As mentioned in the Conclusion, additional parallelism can be combined with our BFS search to further improve the time and make it comparable to beam search. However, even this version of the algorithm can be ``embarrassingly'' parallelized across examples to improve efficiency.

\paragraph{Descriptive Statistics}
We randomly sample 100 data instances from the validation set for each dataset, and they are used by all methods. When sampling is needed, we take 1,000 samples for each data instance, so all the metrics are reported on 100,000 translations/summaries for one dataset.

\section{Risks}

By generating additional outputs from a generation model, we may cause a system to produce outputs that are biased, factually inaccurate, or contain hallucinations. However, all of these risks are present in the raw text generation models as well. Moreover, because we present many options, we believe our approach more appropriately reflects the model's uncertainty over its output, and may have a part to play in mitigating such risks in systems of the future.

\end{document}